\documentclass[11pt,a4paper]{article}
\usepackage[hyperref]{naaclhlt2018}
\usepackage{times}
\usepackage{latexsym}

\usepackage{url}

\usepackage{helvet}  
\usepackage{courier}  
\usepackage{url}  
\usepackage{graphicx}  

\usepackage{algorithm}
\usepackage[noend]{algpseudocode}

\usepackage{eufrak}
\usepackage[font=small]{caption}
\usepackage[font=small]{subcaption}
\usepackage{amsmath}
\usepackage{pgfplots}
\usepackage{tikz}
\usepackage{multirow}
\include{plots}
\definecolor{bblue}{HTML}{4F81BD}
\definecolor{rred}{HTML}{C0504D}
\definecolor{ggreen}{HTML}{9BBB59}
\definecolor{ppurple}{HTML}{9F4C7C}
\definecolor{Dark scarlet}{HTML}{560319}
\definecolor{Forest green}{HTML}{1E4D2B}

\DeclareCaptionFont{10pt}{\fontsize{10pt}{12pt}\selectfont}
\aclfinalcopy 

\title{
Deep Temporal-Recurrent-Replicated-Softmax for \\Topical Trends over Time}
\author{AAAI Press\\
Association for the Advancement of Artificial Intelligence\\
2275 East Bayshore Road, Suite 160\\
Palo Alto, California 94303\\
}

\newcommand*{\affaddr}[1]{#1} 
\newcommand*{\affmark}[1][*]{\textsuperscript{#1}}

\author{Pankaj Gupta\affmark[1,2], Subburam Rajaram\affmark[1], Hinrich Sch\"{u}tze\affmark[2], Bernt Andrassy\affmark[1] \\ 
 \affaddr{\affmark[1]Corporate Technology, Machine-Intelligence (MIC-DE), Siemens AG  Munich, Germany}\\
  \affaddr{\affmark[2]CIS, University of Munich (LMU) Munich, Germany} \\
  {\tt {\{pankaj.gupta, subburam.rajaram, bernt.andrassy\}}@siemens.com}\\
  {\tt pankaj.gupta@campus.lmu.de |  inquiries@cislmu.org}
}

\date{}

\begin{document}
\maketitle

\begin{abstract}
Dynamic topic modeling facilitates the identification of topical trends over time in temporal collections of unstructured documents. 
We introduce a novel unsupervised
neural dynamic topic model 
named as Recurrent Neural Network-Replicated Softmax
Model (RNN-RSM), where the discovered topics at each
time influence the topic discovery in the subsequent
time steps. We account for the temporal ordering of documents
by explicitly modeling a joint distribution of latent topical dependencies over time, using distributional estimators with temporal recurrent connections.
Applying RNN-RSM to 19 years of articles on NLP research, we demonstrate that compared to state-of-the art topic models, RNN-RSM
shows better generalization, topic interpretation,
evolution and trends.
We also introduce a metric (named as SPAN) to quantify the capability of dynamic topic model to capture word evolution in topics over time.
\end{abstract}

\section{Introduction}
Topic Detection and Tracking 
\cite{Allan:82} is an important area of natural language processing to find 
topically related ideas that evolve over time in a sequence of text collections  
and exhibit temporal relationships. 
The temporal aspects of these collections can present valuable insight into 
the topical structure of the collections 
and can be quantified by modeling the dynamics of the underlying topics discovered over time.

\begin{figure}[t]
\begin{minipage}[t]{0.49\linewidth}
    \includegraphics[width=\linewidth]{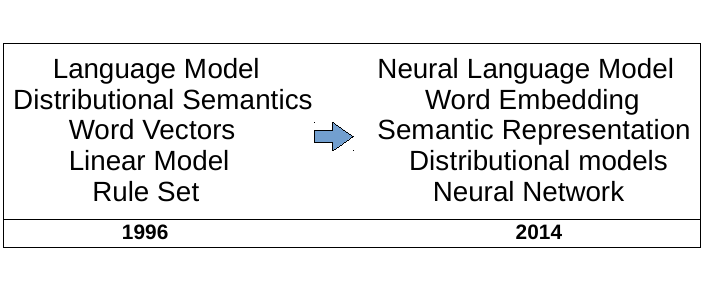}
\end{minipage}
\hfill%
\begin{minipage}[t]{0.49\linewidth}
    \includegraphics[width=\linewidth]{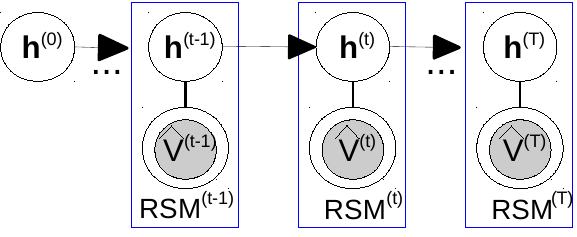}
\end{minipage}
\caption{(Left): Word Usage over time for Topic ({\it Word Representation}) in scholarly articles. 
(Right): RSM-based dynamic topic model with explicit temporal topic dependence}
\label{fig:motivationexampleANDhighlevelRNNRSM}
\end{figure}

{\bf Problem Statement}: We aim to generate temporal topical trends or automatic overview timelines of 
topics  for a time sequence collection of documents. 
This involves the following three tasks in dynamic topic analysis: 
{\bf (1)} 
{\it Topic Structure Detection} (TSD):  
Identifying main topics in the document collection. 
{\bf (2)} 
{\it Topic Evolution Detection} (TED):  
Detecting the emergence of a new topic and recognizing how it grows or decays over time \cite{Allan:83}.    
{\bf (3)} 
{\it Temporal Topic Characterization} (TTC): Identifying the characteristics for each of the main topics 
in order to track the words' usage ({\it keyword trends}) for a topic over time i.e. {\it topical trend analysis for word evolution} (Fig \ref{fig:motivationexampleANDhighlevelRNNRSM}, Left). 

Probabilistic static topic models, such as  Latent Dirichlet Allocation (LDA) \cite{Blei:81} and its variants \cite{xue:82,hall:82,Gol:82} 
have been investigated  to examine the emergence of topics from historical documents. 
Another variant  known as Replicated Softmax (RSM) \cite{Rus:81} has  
demonstrated 
better generalization in log-probability and retrieval, compared to LDA. 
Prior works \cite{Iwata:82, Malinici:82, Saha:82, Schein:82} have investigated Bayesian modeling of topics in time-stamped documents.    
Particularly, \citeauthor{blei:83}  \shortcite{blei:83}  developed a LDA based dynamic topic model (DTM) 
to capture the evolution of topics in a time sequence collection of documents; 
however they do not capture explicitly the topic popularity and usage of specific terms over time.
We propose a family of probabilistic time series models with distributional estimators 
to explicitly model the dynamics of the underlying topics, introducing temporal 
latent topic dependencies (Fig~\ref{fig:motivationexampleANDhighlevelRNNRSM}, Right).

\begin{figure*}[htp]
  \centering
  \includegraphics[width=0.82\linewidth]{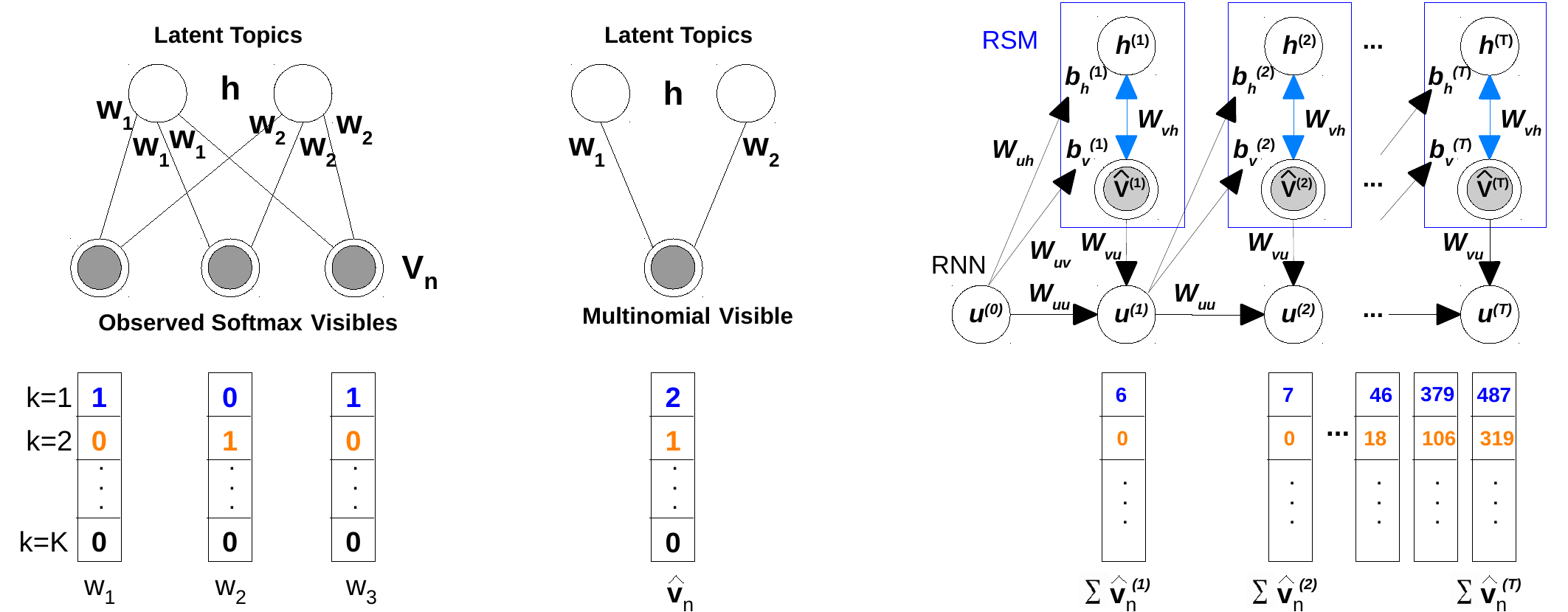}
   \label{fig:RNNRSM}
\caption{(Left): RSM for a document ${\bf V}_{n}$ of $D_{n}$=$3$ words ($w$). 
The bottom layer represents the softmax visible units, 
that share the same set of weights connected to binary hidden units {\bf h}. 
({Middle}): Interpretation of RSM in which $D_{n}$ softmax units with identical
weights are replaced by a single multinomial unit, sampled $D_{n}$ times. 
({Right}): Graphical structure of  2-layered {\bf RNN-RSM}, unfolded in time. 
Single and double headed arrows represent 
deterministic 
and stochastic-
symmetric connections
, respectively. $\widehat{\bf V}^{(t)}$ and ${\bf h}^{(t)}$  
are binary visible and hidden layers of RSM for a document collection at time, $t$. 
${\bf u}$ is RNN hidden layer. $k$: dictionary index for a word $w$}
\label{fig:RNNRSM}
\end{figure*}
To model temporal dependencies in high dimensional sequences, such as  polyphonic music, the temporal stack of RBMs \cite{Smolensky:82,hin:82} has been 
investigated 
to model complex distributions. 
The Temporal RBM \cite{taylorhinton:82,Sutskeverhinton:82}, Recurrent Temporal RBM (RTRBM) \cite{Sutskeverhintontaylor:82} and 
RNN-RBM \cite{nic:82} show success in modeling the temporal dependencies in such symbolic
sequences. 
In addition, RNNs \cite{Gupta:89,Gupta:87,Gupta:88,Gupta:90} have been recognized for sentence modeling in natural language tasks. 
We aspire to build neural dynamic topic model called RNN-RSM to model document collections over time 
and learn temporal topic correlations.

We consider RSM for TSD and introduce the explicit latent topical dependencies for  TED and TTC tasks.
Fig~\ref{fig:motivationexampleANDhighlevelRNNRSM} illustrates our {\it motivation}, where temporal ordering in document collection $\widehat{\bf V}^{(t)}$ at each time step 
$t$, is modeled by conditioning the latent topic 
${\bf h}^{(t)}$ on the sequence history of 
latent topics ${\bf h}^{(0)}$, ..., ${\bf h}^{(t-1)}$, accumulated with temporal lag. 
Each RSM discovers latent topics, where the introduction of a bias term in each RSM 
via the time-feedback latent topic dependencies enables to explicitly model  
topic evolution and specific topic term usage over time. 
The temporal connections and RSM biases allow to convey topical information and model 
relation among the words, in order to deeply analyze the dynamics of the underlying topics. 
We demonstrate the applicability of  proposed {\bf RNN-RSM} by analyzing 19 years of scientific articles from NLP research.

The {\it contributions} in this work are:\\
{\bf (1)} Introduce an unsupervised neural dynamic topic model based on 
recurrent neural network and RSMs, named as RNN-RSM to explicitly model discovered latent topics  (evolution) and word relations (topic characterization) over time.\\
{\bf (2)} Demonstrate better generalization (log-probability and time stamp prediction), topic interpretation (coherence), evolution and characterization, 
compared to the state-of-the-art.\\
{\bf (3)} It is the first work in dynamic topic modeling using undirected stochastic graphical models and deterministic recurrent neural network 
to model collections of different-sized documents over time, within the generative and neural network framework. 
The code and data are available at {\tt https://github.com/pgcool/RNN-RSM}.
\section{The RNN-RSM model}\label{sect:pdf}
RSM (Fig \ref{fig:RNNRSM}, Left) models are a family of different-sized Restricted Boltzmann Machines (RBMs) \cite{Welling:85,Hauptmann:85,Gupta:86,Gupta:91} 
that models {\it word counts} by 
sharing the same parameters with multinomial distribution over the observable i.e. it can be interpreted as a single multinomial unit (Fig \ref{fig:RNNRSM}, Middle) sampled as many times as the document size. 
This facilitates in dealing with the documents of different lengths.

The proposed RNN-RSM model (Fig~\ref{fig:RNNRSM}, Right) is a sequence of conditional RSMs\footnote{Notations: 
$\widehat{\bf U}$=$\{{\bf U}_n\}_{n=1}^{N}$; {\bf U}:2D-Matrix; {\bf l}:vector; {\bf U}/{\bf l}:{\bf U}pper/{\bf l}ower-case; Scalars in unbold} 
such that at any time step $t$, 
the RSM's bias parameters ${\bf b_{v}}^{(t)}$ and ${\bf b_{h}}^{(t)}$  
depend on the output of a deterministic RNN with hidden layer ${\bf u}^{(t-1)}$ in the previous time step, $t-1$. 
Similar to RNN-RBM \cite{nic:82}, we constrain RNN hidden units (${\bf u}^{(t)}$) to convey temporal information, 
while RSM hidden units (${\bf  h}^{(t)}$) to model conditional distributions.
Therefore, parameters (${\bf b_{v}}^{(t)}$, $ {\bf b_{h}}^{(t)}$) 
are time-dependent 
on the sequence history at time $t$ (via a series of conditional RSMs)
denoted by {$\Theta^{(t)} \equiv \{\widehat{\bf V}^{(\tau)}, {\bf u}^{(\tau)}| \tau<t\}$}, 
that captures temporal dependencies. The RNN-RSM is defined by its joint probability distribution:\vspace*{-0.3cm}

{\small\begin{equation*}\label{eq:(1)}
P(\widehat{\bf \mathfrak{V}}, {\bf H}) =  P(\{ \widehat{\bf V}^{(t)}, {\bf h}^{(t)}\}_{t=1}^{T})  = \prod_{t=1}^{T} P(\widehat{\bf V}^{(t)}, {\bf h}^{(t)} | \Theta^{(t)})   
\end{equation*}}
where {$\widehat{\bf \mathfrak{V}}$ = [$\widehat{\bf V}^{(1)}, ...\widehat{\bf V}^{(T)}$]} and 
{${\bf H}$ = [${\bf h}^{(1)}, ...{{\bf h}^{(T)}}$]}. 
Each {${\bf h}^{(t)}$ $\in \{0,1\}^{F}$} be a binary stochastic hidden topic vector with size $F$ and 
{$\widehat{\bf V}^{(t)} = \{{\bf V}_{n}^{(t)}\}_{n=1}^{N^{(t)}}$} 
be a collection of $N$ documents at time step $t$. 
Let {${\bf V}_{n}^{(t)}$} be a {$K \times D_{n}^{(t)}$} observed binary matrix of the $n^{th}$ 
document in the collection 
where, {$D_{n}^{(t)}$} is the document size and $K$  is the dictionary size over all the time steps. 
The conditional distribution (for each unit in hidden or visible) in each RSM at time step, 
is given by softmax and logistic functions:\vspace*{-0.3cm}

{\small\begin{align*}
\begin{split}
P(v_{n, i}^{k, (t)}=1| {\bf h}_{n}^{(t)}) & = \frac{{\exp}( {b_{v,i}}^{k, (t)} + \sum_{j=1}^{F}h_{n, j}^{(t)}{W}_{ij}^{k})}
{\sum_{q=1}^{{K}}{\exp}({b_{v,i}}^{q, (t)} + \sum_{j=1}^F h_{n, j}^{(t)}{W}^{q}_{ij})} \\
P(h_{n, j}^{(t)} = 1| &{\bf V}_{n}^{(t)}) =  \sigma (b_{h,j}^{(t)}+ \sum_{i=1}^{D_{n}^{(t)}}\sum_{k=1}^{K} v_{n, i}^{k, (t)}W_{ij}^{k})
\end{split}
\end{align*}}
where  
{\small $P(v_{n, i}^{k, (t)} = 1| {\bf h}_{n}^{(t)})$} and {\small $P(h_{n, j}^{(t)} = 1| {\bf V}_{n}^{(t)})$} are 
conditional distributions for $i^{th}$ visible $v_{n, i}$ and $j^{th}$ hidden unit $h_{n, j}$ for the $n^{th}$ document at $t$. 
$W^{k}_{ij}$ is a symmetric interaction term between $i$ that takes on value $k$ and $j$. 
$v^{k, (t)}_{n}$ is sampled $D_{n}^{(t)}$ times with identical weights connected to binary 
hidden units, resulting in multinomial visibles, therefore the name {\it Replicated Softmax}.  
The conditionals across layers are factorized as: {\small $P({\bf V}^{(t)}_{n}|{\bf h}_{n}^{(t)}) = \prod_{i=1}^{{D}_{n}^{(t)}} P({\bf v}^{(t)}_{n, i}| {\bf h}_{n}^{(t)})$};  
{\small $P({\bf h}_{n}^{(t)} | {\bf V}^{(t)}_{n}) = \prod_{j} P(h_{n, j}^{(t)} |  {\bf V}^{(t)}_{n})$}. 

Since biases of RSM depend on the output of RNN at previous time steps, that allows to propagate the estimated gradient at each RSM 
backward through time (BPTT). The {\it RSM biases} 
and RNN hidden state ${\bf u}^{(t)}$ at each time step $t$ are given by- \vspace*{-0.3cm}

\begin{algorithm}[t]
\caption{\textit{Training RNN-RSM with BPTT}}\label{trainingRNNRSM}
\small{ 
\begin{algorithmic}[1]
\Statex \textbf{Input}: \textit{Observed visibles}, $\widehat{\bf \mathfrak{V}}$ = \{$\widehat{\bf V}^{(0)}, \widehat{\bf V}^{(1)},..., \widehat{\bf V}^{(t)},..., \widehat{\bf V}^{(T)}$\}
\Statex \textbf{RNN-RSM Parameters}: $\bf {\theta}$ = \{${\bf W_{uh}}$, ${\bf W_{vh}}$, ${\bf W_{uv}}$, ${\bf W_{vu}}$, 
${\bf W_{uu}}$, ${\bf b_{v}}$,   ${\bf b_{u}}$, ${\bf b_{h}}$, ${\bf b_{v}}^{(t)}$, ${\bf b_{h}}^{(t)}$, ${\bf u}^{(0)}$\}
\State Propagate 
${\bf u}^{(t)}$ in RNN portion of the graph using eq \ref{eq:(6)}.
\State Compute ${\bf b_{v}}^{(t)}$ and ${\bf b_{h}}^{(t)}$ 
using eq \ref{eq:(4)}.
\State Generate negatives ${\bf V}^{(t)*}$ using k-step  Gibbs sampling.
\State Estimate the gradient of the cost $C$ 
w.r.t. parameters of RSM  
${\bf W_{vh}}$,  ${\bf b_{v}}^{(t)}$ and ${\bf b_{h}}^{(t)}$  using eq \ref{eq:(12)}.
\State Compute gradients (eq \ref{eq:gradwrtRNN}) w.r.t. RNN connections (${\bf W_{uh}}$, ${\bf W_{uv}},  {\bf W_{uu}},  {\bf W_{vu}}, {\bf u}^{0}$) 
 and biases (${\bf b_{v}}$, ${\bf b_{h}}$, ${\bf b_{u}}$).
\State \textbf{Goto} \emph{step 1} until stopping$\_$criteria  (early stopping or maximum iterations reached)
\end{algorithmic}}
\end{algorithm}

{\small 
\begin{align}
\begin{split}\label{eq:(4)}
{\bf b_{v}}^{(t)}  = {\bf b_v} + & {\bf W_{uv}} {\bf u}^{(t-1)} \\
{\bf b_{h}}^{(t)}  = {\bf b_h} + & {\bf W_{uh}} {\bf u}^{(t-1)}
\end{split}\\
\begin{split}\label{eq:(6)}
{\bf u}^{(t)} = \tanh ( {\bf b_u} +  {\bf W_{uu}} & {\bf u}^{(t-1)} +   {\bf W_{vu}} \sum_{n=1}^{N^{(t)}} \hat{\bf v}_{n}^{(t)} )
\end{split}
\end{align}}
where ${\bf W_{uv}}$, ${\bf{W}_{uh}}$  and ${\bf W_{vu}}$ are weights connecting RNN and RSM portions (Figure \ref{fig:RNNRSM}). 
${\bf  b_{u}}$ is the bias of ${\bf u}$ and ${\bf W_{uu}}$ is the weight between RNN hidden units. 
$\hat{\bf v}_{n}^{(t)}$ is a vector of $\hat v_{n}^{k}$ (denotes the count for the $k^{th}$ word in $n^{th}$ document).
$\sum_{n=1}^{N^{(t)}} \hat{\bf v}_{n}^{(t)}$ refers to the sum of observed vectors across documents at time step $t$ where each document is represented as-\vspace*{-0.3cm}
\begin{equation}\label{eq:sumdocs}
\hat{\bf v}_{n}^{(t)} =  [\{ \hat{v}_{n}^{k, (t)}\}_{k=1}^{K}]  \ \ \mbox{and} \ \ \hat v_{n}^{k, (t)} = \sum_{i=1}^{D_{n}^{(t)}}v_{n, i}^{k, (t)} 
\end{equation}
where $v^{k, (t)}_{n, i}$=1 if visible unit $i$ takes on $k^{th}$ value.

In each RSM, a separate RBM is created for each document in the collection at time step $t$ with $D_{n}^{(t)}$ softmax units, 
where $D_{n}^{(t)}$ is the count of words in the $n^{th}$ document. 
Consider a document of $D_{n}^{(t)}$ words, the {\it energy} of the state $\{{\bf V}_{n}^{(t)}, {\bf h}_{n}^{(t)}\}$ at time step, $t$ is given by-\vspace*{-0.4cm}

{\small \begin{align*}
\begin{split}
E({\bf V}_{n}^{(t)}, {\bf h}_{n}^{(t)})   =&  -\sum_{j=1}^{F}  \sum_{k=1}^{{K}}  h_{n, j}^{(t)} {W}^{k}_{j}  \hat v^{k, (t)}_{n}\\
 & -   \sum_{k=1}^\mathrm{K}  \hat v^{k, (t)}_{n} b_{v}^{k} -  D_{n}^{(t)}\sum_{j=1}^{F} b_{h, j} h_{n, j}^{(t)}
\end{split}
\end{align*}} \vspace*{-0.3cm}

\begin{table*}[t]
\centering
\def\arraystretch{1.0}
\resizebox{\textwidth}{!}{
\begin{tabular}{c|rrrrrrrrrrrrrrrrrrr|c}
{\bf Year}  & 1996 & 1997 & 1998 & 1999 & 2000 & 2001 & 2002 & 2003 & 2004 & 2005 & 2006 & 2007 & 2008 & 2009 & 2010 & 2011 & 2012 & 2013 & 2014 & {\bf Total} \\
\hline
{\bf ACL}& 58   & 73   & 250  & 83  & 79   & 70   & 177  & 112  & 134  & 134  & 307  & 204  & 214  & 243  & 270  & 349  & 227  & 398  & 331 & 3713
\\
{\bf EMNLP}& 15   & 24   & 15  & 36  & 29   & 21   & 42  & 29  & 58  & 28  & 75  & 132  & 115  & 164  & 125  & 149  & 140  & 206  & 228 & 1756
\\ \hline
{\bf ACL+EMNLP}& 73   & 97   & 265  & 119  & 108   & 91   & 219  & 141  & 192  & 162  & 382  & 336  & 329  & 407  & 395  & 498  & 367  & 604  & 559 & {\bf 5469}
\end{tabular}}
\caption{Number of papers from ACL and EMNLP conferences over the years}
\label{paperscount}
\end{table*}

Observe that the bias terms on hidden units are scaled up by document length to allow hidden units to stabilize 
when dealing with different-sized documents.
The corresponding energy-probability relation in the energy-based model is-

{\small
\begin{equation}\label{eq:(8)}
P({\bf V}_{n}^{(t)}) = \frac{1}{Z_{n}^{(t)}}\sum_{{\bf h}_{n}^{(t)}} \exp(-E({\bf V}_{n}^{(t)}, {\bf h}_{n}^{(t)}))   
\end{equation}}
where $Z_{n}^{(t)}$ = $\sum_{{\bf  V}_{n}^{(t)}} \sum_{{\bf h}_{n}^{(t)}}\exp(-E({\bf V}_{n}^{(t)}, {\bf h}_{n}^{(t)}))$ is  the normalization constant.
The lower bound on the log likelihood of the data takes the form:\vspace*{-0.3cm}

{\small \begin{eqnarray*}
\ \ \ \ln P({\bf V}_{n}^{(t)})  \ge \sum_{{\bf h}^{(t)}} Q({\bf h}_{n}^{(t)} | {\bf V}_{n}^{(t)}) \ln P({\bf V}_{n}^{(t)}, {\bf h}_{n}^{(t)}) + H(Q) \\
=  \ln P({\bf V}_{n}^{(t)}) - KL[ Q({\bf h}_{n}^{(t)} | {\bf V}_{n}^{(t)}) ||  P({\bf h}_{n}^{(t)} | {\bf V}_{n}^{(t)})] 
\end{eqnarray*}}
where $H(\cdot)$ is the entropy and $Q$ is the approximating posterior. 
Similar to Deep Belief Networks \cite{hinton:82}, adding an extra layer improves lower bound on the log probability of data, 
we introduce the extra layer via RSM biases that
propagates the prior via RNN connections. The dependence analogy follows-\vspace*{.3cm}

{\small $ E({\bf V}_{n}^{(t)}, {\bf h}_{n}^{(t)})  \propto  \frac{1}{{\bf b_{v}}^{(t)}}   \ \mbox{and}  \  E({\bf V}_{n}^{(t)}, {\bf h}_{n}^{(t)})  \propto \frac{1}{{\bf b_{h}}^{(t)}}$ \vspace*{0.25cm} \\ \vspace*{0.3cm}
$\ln P({\bf V}_{n}^{(t)})  \propto  \frac{1}{E({\bf V}_{n}^{(t)}, {\bf h}_{n}^{(t)})} ;  \ln P(\widehat{\bf V}_{n}^{(t)}) \propto  \ln P(\{\widehat{\bf V}_{n}^{\tau}\}_{\tau <t})$}

Observe that the prior is seen as the deterministic hidden representation of latent topics and injected into each hidden state of RSMs, 
that enables the likelihood of the data to model complex temporal densities 
i.e. heteroscedasticity in document collections ($\widehat{\bf \mathfrak{V}}$) and temporal topics (${\bf H}$). 

{\bf Gradient Approximations:} The {\it cost} in RNN-RSM is:
$\ \ C = \sum_{t=1}^{T} C_{t} \equiv \sum_{t=1}^{T}  - \ln P(\widehat{\bf V}^{(t)})$

Due to intractable $Z$, the gradient of cost at time step $t$  
w.r.t. (with respect to) RSM parameters are approximated by 
k-step Contrastive Divergence (CD)  \cite{hin:82}. 
The gradient of the negative log-likelihood of a document collection $\{{\bf V}_{n}^{(t)}\}_{n=1}^{N^{(t)}}$ w.r.t. RSM parameter ${\bf W_{vh}}$,
{\begin{equation*}
\begin{split} \label{eq:(10)}
&\frac{1}{N^{(t)}}  \sum_{n=1}^{N^{(t)}}  \frac{\partial (- \ln P({\bf V}_{n}^{(t)}))}{\partial {\bf W_{vh}}}  \\
& = \frac{1}{N^{(t)}} \sum_{n=1}^{N^{(t)}} \frac{\partial \mathfrak{F}({\bf V}_{n}^{(t)})}{\partial {\bf W_{vh}}} - \frac{\partial  (- \ln Z_{n}^{(t)})}{\partial {\bf W_{vh}}} \\
& = \underbrace{\textsf{E}_{P_{data}}[\frac{\partial \mathfrak{F}({\bf V}_{n}^{(t)})}{\partial {\bf W_{vh}}}]}_\text{data-dependent expectation}  -  \underbrace{\textsf{E}_{P_{model}}[\frac{\partial \mathfrak{F}({\bf V}_{n}^{(t)})}{\partial {\bf W_{vh}}}]}_\text{model's expectation}  \\
& \simeq  \frac{1}{N^{(t)}} \sum_{n=1}^{N^{(t)}} \frac{\partial \mathfrak{F}({\bf V}_{n}^{(t)})}{\partial {\bf W_{vh}}} - \frac{\partial \mathfrak{F}({\bf V}_{n}^{(t)*})}{\partial {\bf W_{vh}}}
\end{split}
\end{equation*}}
The second term is estimated by negative samples ${\bf V}_{n}^{(t)*}$
obtained from k-step Gibbs chain starting at ${\bf V}_{n}^{(t)}$ samples. 
{$P_{data}(\widehat{\bf V}^{(t)}, {\bf h}^{(t)}) = P({\bf h}^{(t)} | \widehat{\bf V}^{(t)}) P_{data}(\widehat{\bf V}^{(t)})$} and 
{$P_{data}(\widehat{\bf V}^{(t)}) = \frac{1}{N^{(t)}}\sum_{n}^{N^{(t)}} \delta(\widehat{\bf V}^{(t)} - {\bf V}_{n}^{(t)})$} is 
the empirical distribution on the observable. 
{$P_{model}({\bf V}_{n}^{(t)*}, {\bf h}_{n}^{(t)})$} is defined in eq.~\ref{eq:(8)}.
The free energy $\mathfrak{F}({\bf V}_{n}^{(t)})$ is related to normalized probability of 
{${\bf V}_{n}^{(t)}$ as $P({\bf V}_{n}^{(t)})$ $\equiv$ $\exp^{-\mathfrak{F}({\bf V}_{n}^{(t)})}/Z_{n}^{(t)}$} 
and as follows-
{\small 
\begin{align*}
\begin{split}
\mathfrak{F}({\bf V}_{n}^{(t)})  = - \sum_{k=1}^{K} & \hat{v}_{n}^{k, (t)}b_{v}^{k} -  \sum_{j=1}^{F}  \log(1    +  \\
& \exp(D^{(t)}_{n} b_{h, j} +  \sum_{k=1}^{K}  \hat{v}_{n}^{k, (t)} W^{k}_{j}))
\end{split}
\end{align*}}
Gradient approximations w.r.t. RSM parameters, 
\begin{align}
\begin{split}\label{eq:(12)}
\frac{\partial C_{t}}{\partial {\bf b_{v}}^{(t)}} &   \simeq   \sum_{n=1}^{N^{(t)}} {\bf \hat{v}}_{n}^{(t)*} - {\bf \hat{v}}_{n}^{(t)}   \\
\; \; \ \ \ \ \ \frac{\partial C_{t}}{\partial {\bf b_{h}}^{(t)}}  \simeq  \sum_{n=1}^{N^{(t)}} \sigma & ({\bf W_{vh}}  {\bf \hat{v}}_{n}^{(t)*} 
-  D^{(t)}_{n} {\bf b_{h}}^{(t)}) \\
 -  \sigma & ({\bf W_{vh}} {\bf \hat{v}}_{n}^{(t)} - D^{(t)}_{n} {\bf b_{h}}^{(t)})  \\
\frac{\partial C_{t}}{\partial {\bf W_{vh}}} \simeq  \sum_{t=1}^{T}  \sum_{n=1}^{N^{(t)}} \sigma (&{\bf W_{vh}} {\bf \hat{v}}_{n}^{(t)*} 
 -  D^{(t)}_{n} {\bf b_{h}}^{(t)})\\ 
{\bf \hat{v}}_{n}^{(t)*T} - \sigma(&{\bf W_{vh}}  {\bf \hat{v}}_{n}^{(t)} -  D^{(t)}_{n} {\bf b_{h}}^{(t)}) {\bf \hat{v}}_{n}^{(t)T}
\end{split}
\end{align}
The estimated gradients w.r.t. RSM biases are back-propagated via hidden-to-bias parameters (eq \ref{eq:(4)}) to compute gradients w.r.t. RNN 
connections 
({${\bf W_{uh}}$, ${\bf W_{uv}}$,  ${\bf W_{vu}}$} and {${\bf W_{uu}}$}) and biases ({${\bf b_{h}}$, ${\bf b_{v}}$}  and  ${\bf b_{u}}$).  
\begin{align}
\begin{split} \label{eq:gradwrtRNN}
\frac{\partial C}{\partial {\bf W_{uh}}}    & =  \sum_{t=1}^{T} \frac{\partial C_{t}}{\partial {\bf b_{h}}^{(t)}} {\bf u}^{(t-1)T} \\
\frac{\partial C}{\partial {\bf W_{uv}}}   & =   \sum_{t=1}^{T} \frac{\partial C_{t}}{\partial {\bf b_{v}}^{(t)}} {\bf u}^{(t-1)T} \\
\frac{\partial C}{\partial {\bf W_{vu}}}   = &  \sum_{t=1}^{T} \frac{\partial C_{t}}{\partial {\bf u}^{(t)}} {\bf u}^{(t)} (1- {\bf u}^{(t)}) \sum_{n=1}^{N^{(t)}}  \hat{\bf v}_{n}^{(t)T}\\
\frac{\partial C}{\partial {\bf b_{h}}}   = &   \sum_{t=1}^{T}   \frac{\partial C_{t}}{\partial {\bf b_{h}}^{(t)}}  \ \mbox{and}\  
\frac{\partial C}{\partial {\bf b_{v}}}  = \sum_{t=1}^{T} \frac{\partial C_{t}}{\partial {\bf b_{v}}^{(t)}} \\
\frac{\partial C}{\partial {\bf b_u}} & = \sum_{t=1}^{T} \frac{\partial C_{t}}{\partial {\bf u}^{(t)}} {\bf u}^{(t)} ({1 - \bf u}^{(t)})\\
\frac{\partial C}{\partial {\bf W_{uu}}} & = \sum_{t=1}^{T} \frac{\partial C_{t}}{\partial {\bf u}^{(t)}} {\bf u}^{(t)} ({1 - \bf u}^{(t)}) {\bf u}^{(t-1)T}
\end{split}
\end{align}
For the single-layer RNN-RSM, the BPTT recurrence relation for $ 0 \le t <  T $  is given by-
\begin{align*}
\begin{split}
\frac{\partial C_{t}}{\partial {\bf u}^{(t)}} =  {\bf W_{uu}} \frac{\partial C_{t+1}}{\partial {\bf u}^{(t+1)}} {\bf u}^{(t+1)} (1 - {\bf u}^{(t+1)}) \\
+ {\bf W_{uh}} \frac{\partial C_{t+1}}{\partial {\bf b_{h}}^{(t+1)}} + {\bf W_{uv}} \frac{\partial C_{t+1}}{\partial {\bf b_{v}}^{(t+1)}}
\end{split}
\end{align*}
where ${\bf u}^{(0)}$ being a parameter and $ \frac{\partial C_{T}}{\partial {\bf u}^{(T)}}= 0$. 

See {\it Training RNN-RSM with BPTT} in Algo \ref{trainingRNNRSM}.

\begin{figure*}[t] 
\centering
\begin{subfigure}{0.25\textwidth}
\centering
\begin{tikzpicture}[scale=0.45,trim axis left, trim axis right][baseline]
\begin{axis}[
    xlabel={Years},
    ylabel={Cosine Similarity},
    xmin=0, xmax=19,
    ymin=0.0, ymax=0.8,
    xtick={0,1,2,3,4,5,6,7,8,9,10,11,12,13,14,15,16,17,18,19},
    ytick={0,0.1,0.2,0.3,0.4,0.5,0.6,0.7},
    xticklabels={96,97,98,99,00,01,02,03,04,05,06,07,08,09,10,11,12,13,14},
    x tick label style={rotate=90,anchor=east},
    legend pos=north west,
    ymajorgrids=true,
    grid style=dashed,
]
\addplot[
    color=blue,
    mark=square,
    ]
    plot coordinates {
    (0,0.052704627669473036)
    (1,0.052704627669473036)
    (2,0.052704627669473036)
    (3,0.052704627669473036)
    (4,0.052704627669473036)
    (5,0.052704627669473036)
    (6,0.158113883008419)
    (7,0.158113883008419)
    (8,0.21081851067789192)
    (9,0.21081851067789192)
    (10,0.21081851067789192)
    (11,0.26352313834736496)
    (12,0.26352313834736496)
    (13,0.26352313834736496)
    (14,0.26352313834736496)
    (15,0.26352313834736496)
    (16, 0.2635231383473649)
    (17,0.2635231383473649)
    (18,0.2635231383473649)
    };

\addlegendentry{RNN-RSM}

\addplot[
	color=red,
	mark=*,
	]
	plot coordinates {
    (0,0.052704627669473036)
    (1,0.052704627669473036)
    (2,0.052704627669473036)
    (3,0.052704627669473036)
    (4,0.0)
    (5,0.052704627669473036)
    (6,0.0)
    (7,0.052704627669473036)
    (8,0.052704627669473036)
    (9,0.052704627669473036)
    (10,0.052704627669473036)
    (11,0.21081851067789192)
    (12,0.26352313834736496)
    (13,0.21081851067789192)
    (14,0.158113883008419)
    (15,0.052704627669473036)
    (16,0.21081851067789192)
    (17,0.10540925533894596)
    (18,0.21081851067789192)
	};
\addlegendentry{RSM}

\addplot[
    color=Forest green,
    mark=x,
    ]
    plot coordinates {
    (0,0.0)
    (1,0.0)
    (2,0.0)
    (3,0.054232614454664096)
    (4,0.0)
    (5,0.0)
    (6,0.0)
    (7,0.054232614454664096)
    (8,0.10846522890932808)
    (9,0.054232614454664096)
    (10,0.16269784336399207)
    (11,0.16269784336399207)
    (12,0.21693045781865616)
    (13,0.21693045781865616)
    (14,0.27116307227332026)
    (15,0.27116307227332026)
    (16,0.27116307227332026)
    (17,0.21693045781865616)
    (18,0.27116307227332026)
    };

\addlegendentry{LDA}

\addplot[
    color=gray,
    mark=diamond,
    ]
    plot coordinates {
    (0,0.0)
    (1,0.0)
    (2,0.054232614454664096)
    (3,0.054232614454664096)
    (4,0.054232614454664096)
    (5,0.10846522890932808)
    (6,0.10846522890932808)
    (7,0.10846522890932808)
    (8,0.10846522890932808)
    (9,0.16269784336399207)
    (10,0.16269784336399207)
    (11,0.16269784336399207)
    (12,0.16269784336399207)
    (13,0.16269784336399207)
    (14,0.16269784336399207)
    (15,0.21693045781865616)
    (16,0.21693045781865616)
    (17,0.21693045781865616)
    (18,0.21693045781865616)
    };

\addlegendentry{DTM}

\end{axis}
\end{tikzpicture}%
\caption{{Topic}: Sentiment Analysis} \label{Standard cosine similarity Sentiment Analysis.}
\end{subfigure}\hspace*{\fill}%
~%
\begin{subfigure}{0.25\textwidth}
\centering
\begin{tikzpicture}[scale=0.45,trim axis left, trim axis right][baseline]
\begin{axis}[
    xlabel={Years},
    ylabel={Cosine Similarity},
    xmin=0, xmax=19,
    ymin=0.0, ymax=0.8,
    xtick={0,1,2,3,4,5,6,7,8,9,10,11,12,13,14,15,16,17,18,19},
    ytick={0,0.1,0.2,0.3,0.4,0.5,0.6,0.7},
    xticklabels={96,97,98,99,00,01,02,03,04,05,06,07,08,09,10,11,12,13,14},
    x tick label style={rotate=90,anchor=east},
    legend pos=north west,
    ymajorgrids=true,
    grid style=dashed,
]
\addplot[
    color=blue,
    mark=square,
    ]
    plot coordinates {
    (0,0.093658581158169385)
    (1,0.15389675281277304)
    (2,0.2051956704170308)
    (3,0.2051956704170308)
    (4,0.15389675281277304)
    (5,0.1025978352085154)
    (6,0.2051956704170308)
    (7,0.15389675281277304)
    (8,0.2051956704170308)
    (9,0.2051956704170308)
    (10,0.2051956704170308)
    (11,0.25649458802128855)
    (12,0.2051956704170308)
    (13,0.2051956704170308)
    (14,0.25649458802128855)
    (15,0.25649458802128855)
    (16,0.25649458802128855)
    (17,0.25649458802128855)
    (18,0.25649458802128855)
    };
\addlegendentry{RNN-RSM}

\addplot[
	color=red,
	mark=*,
	]
	plot coordinates {
    (0,0.051298917604257754)
    (1,0.051298917604257754)
    (2,0.051298917604257754)
    (3,0.051298917604257754)
    (4,0.051298917604257754)
    (5,0.051298917604257754)
    (6,0.051298917604257754)
    (7,0.051298917604257754)
    (8,0.051298917604257754)
    (9,0.051298917604257754)
    (10,0.051298917604257754)
    (11,0.1025978352085154)
    (12,0.051298917604257754)
    (13,0.15389675281277304)
    (14,0.15389675281277304)
    (15,0.1025978352085154)
    (16,0.051298917604257754)
    (17,0.1025978352085154)
    (18,0.051298917604257754)
	};
\addlegendentry{RSM}

\addplot[
    color=Forest green,
    mark=x,
    ]
    plot coordinates {
    (0,0.051298917604257754)
    (1,0.1025978352085154)
    (2,0.1025978352085154)
    (3,0.051298917604257754)
    (4,0.1025978352085154)
    (5,0.1025978352085154)
    (6,0.1025978352085154)
    (7,0.15389675281277304)
    (8,0.1025978352085154)
    (9,0.1025978352085154)
    (10,0.1025978352085154)
    (11,0.1025978352085154)
    (12,0.1025978352085154)
    (13,0.1025978352085154)
    (14,0.1025978352085154)
    (15,0.1025978352085154)
    (16,0.15389675281277304)
    (17,0.15389675281277304)
    (18,0.25649458802128855)
    };

\addlegendentry{LDA}

\addplot[
    color=gray,
    mark=diamond,
    ]
    plot coordinates {
    (0,0.1025978352085154)
    (1,0.1025978352085154)
    (2,0.1025978352085154)
    (3,0.1025978352085154)
    (4,0.1025978352085154)
    (5,0.1025978352085154)
    (6,0.1025978352085154)
    (7,0.1025978352085154)
    (8,0.1025978352085154)
    (9,0.1025978352085154)
    (10,0.1025978352085154)
    (11,0.1025978352085154)
    (12,0.1025978352085154)
    (13,0.1025978352085154)
    (14,0.1025978352085154)
    (15,0.1025978352085154)
    (16,0.1025978352085154)
    (17,0.1025978352085154)
    (18,0.1025978352085154)
    };

\addlegendentry{DTM}

\end{axis}
\end{tikzpicture}%
\caption{{Topic}: Word Vector} \label{Standard cosine similarity Word Vector.}
\end{subfigure}\hspace*{\fill}%
~%
\begin{subfigure}{0.25\textwidth}
\centering
\begin{tikzpicture}[scale=0.45,trim axis left, trim axis right][baseline]
\begin{axis}[
    xlabel={Years},
    ylabel={Cosine Similarity},
    xmin=0, xmax=19,
    ymin=0.0, ymax=0.8,
    xtick={0,1,2,3,4,5,6,7,8,9,10,11,12,13,14,15,16,17,18,19},
    ytick={0,0.1,0.2,0.3,0.4,0.5,0.6,0.7},
    xticklabels={96,97,98,99,00,01,02,03,04,05,06,07,08,09,10,11,12,13,14},
    x tick label style={rotate=90,anchor=east},
    legend pos=north west,
    ymajorgrids=true,
    grid style=dashed,
]
\addplot[
    color=blue,
    mark=square,
    ]
    plot coordinates {
    (0,0.21516574145596756)
    (1,0.34426518632954806)
    (2,0.34426518632954806)
    (3,0.3872983346207417)
    (4,0.34426518632954806)
    (5,0.34426518632954806)
    (6,0.34426518632954806)
    (7,0.34426518632954806)
    (8,0.34426518632954806)
    (9,0.34426518632954806)
    (10,0.3872983346207417)
    (11,0.3872983346207417)
    (12,0.3872983346207417)
    (13,0.3872983346207417)
    (14,0.3872983346207417)
    (15,0.3872983346207417)
    (16,0.3872983346207417)
    (17,0.3872983346207417)
    (18,0.3872983346207417)
    };
\addlegendentry{RNN-RSM}

\addplot[
	color=red,
	mark=*,
	]
	plot coordinates {
    (0,0.12909944487358049)
    (1,0.12909944487358049)
    (2,0.08606629658238707)
    (3,0.08606629658238707)
    (4,0.08606629658238707)
    (5,0.12909944487358049)
    (6,0.12909944487358049)
    (7,0.08606629658238707)
    (8,0.043033148291193535)
    (9,0.08606629658238707)
    (10,0.12909944487358049)
    (11,0.2581988897471611)
    (12,0.21516574145596756)
    (13,0.34426518632954806)
    (14,0.043033148291193535)
    (15,0.12909944487358049)
    (16,0.17213259316477403)
    (17,0.08606629658238707)
    (18,0.17213259316477403)
    };
\addlegendentry{RSM}

\addplot[
    color=Forest green,
    mark=x,
    ]
    plot coordinates {
    (0,0.21516574145596756)
    (1,0.17213259316477403)
    (2,0.2581988897471611)
    (3,0.30123203803835463)
    (4,0.21516574145596756)
    (5,0.30123203803835463)
    (6,0.30123203803835463)
    (7,0.2581988897471611)
    (8,0.34426518632954806)
    (9,0.30123203803835463)
    (10,0.34426518632954806)
    (11,0.2581988897471611)
    (12,0.3872983346207417)
    (13,0.34426518632954806)
    (14,0.34426518632954806)
    (15,0.34426518632954806)
    (16,0.30123203803835463)
    (17,0.30123203803835463)
    (18,0.34426518632954806)
    };
\addlegendentry{LDA}

\addplot[
    color=gray,
    mark=diamond,
    ]
    plot coordinates {
    (0,0.17541160386140586)
    (1,0.17541160386140586)
    (2,0.17541160386140586)
    (3,0.17541160386140586)
    (4,0.2192645048267573)
    (5,0.2192645048267573)
    (6,0.2192645048267573)
    (7,0.26311740579210885)
    (8,0.30697030675746029)
    (9,0.30697030675746029)
    (10,0.26311740579210885)
    (11,0.2192645048267573)
    (12,0.2192645048267573)
    (13,0.26311740579210885)
    (14,0.26311740579210885)
    (15,0.26311740579210885)
    (16,0.26311740579210885)
    (17,0.26311740579210885)
    (18,0.26311740579210885)
    };

\addlegendentry{DTM}

\end{axis}
\end{tikzpicture}%
\caption{{Topic}: Dependency Parsing} \label{Standard cosine similarity Dependency Parsing.}
\end{subfigure}\hspace*{\fill}%
~%
\begin{subfigure}{0.24\textwidth}
\centering
\begin{tikzpicture}[scale=0.45,trim axis left, trim axis right][baseline]
\begin{axis}[
    xlabel={Years},
    ylabel={Perplexity (PPL)},
    xmin=0, xmax=18,
    ymin=0.0, ymax=0.9,
    xtick={0,1,2,3,4,5,6,7,8,9,10,11,12,13,14,15,16,17,18},
    xticklabels={96,97,98,99,00,01,02,03,04,05,06,07,08,09,10,11,12,13,14},
    x tick label style={rotate=90,anchor=east},
    legend pos=north west,
    ymajorgrids=true,
    grid style=dashed,
]
\addplot[
    color=blue,
    mark=square,
    ]
    plot coordinates {
    (0, 0.16)
    (1, 0.16)
    (2, 0.18)
    (3, 0.20)
    (4, 0.25)
    (5, 0.19)
    (6, 0.23)
    (7, 0.25)
    (8, 0.19)
    (9, 0.18)
    (10, 0.22)
    (11, 0.10)
    (12, 0.19)
    (13, 0.15)
    (14, 0.23)
    (15, 0.24)
    (16,  0.22)
    (17, 0.20)
    (18, 0.20)
    };

\addlegendentry{RNN-RSM}

\addplot[
	color=gray,
	mark=diamond,
	]
	plot coordinates {
    (0, 0.70)
    (1, 0.69)
    (2, 0.62)
    (3, 0.59)
    (4, 0.19)
    (5, 0.74)
    (6, 0.42)
    (7, 0.53)
    (8, 0.72)
    (9, 0.52)
    (10, 0.63)
    (11, 0.63)
    (12, 0.55)
    (13, 0.57)
    (14, 0.54)
    (15, 0.60)
    (16, 0.54)
    (17, 0.58)
    (18, 0.56)
	};
\addlegendentry{DTM}
\end{axis}
\end{tikzpicture}%
\caption{Perplexity on Unobserved} \label{Perplexity over time unobserved}
\end{subfigure}\hspace*{\fill}

\medskip

\begin{subfigure}{0.25\textwidth}
\centering
\begin{tikzpicture}[scale=0.45,trim axis left, trim axis right][baseline]
\begin{axis}[
    xlabel={Years},
    ylabel={COH (mean) of topics: COH  $\times$ $10^{-2}$},
    xmin=0, xmax=18,
    ymin=14.0, ymax=18.0,
    xtick={0,1,2,3,4,5,6,7,8,9,10,11,12,13,14,15,16,17,18},
    ytick={14.0, 14.5, 15.0, 15.5, 16.0, 16.5, 17.0,17.5,18.0},
    xticklabels={96,97,98,99,00,01,02,03,04,05,06,07,08,09,10,11,12,13,14},
    x tick label style={rotate=90,anchor=east},
    legend pos=north west,
    ymajorgrids=true,
    grid style=dashed,
]
\addplot[
    color=blue,
    mark=square,
    ]
    plot coordinates {
    (0, 15.8)
    (1, 16.3)
    (2, 16.3)
    (3, 16.1)
    (4, 16.1)
    (5, 15.8)
    (6, 16.1)
    (7, 15.9)
    (8, 16.0)
    (9, 15.8)
    (10, 15.9)
    (11, 16.4)
    (12, 16.0)
    (13, 16.4)
    (14, 16.3)
    (15, 16.3)
    (16,  16.7)
    (17, 16.5)
    (18, 16.2)
    };
\addlegendentry{RNN-RSM}
\addplot[
	color=gray,
	mark=diamond,
	]
	plot coordinates {
    (0, 15.0)
    (1, 15.3)
    (2, 15.3)
    (3, 15.3)
    (4, 15.2)
    (5, 15.3)
    (6, 15.7)
    (7, 16.3)
    (8, 15.9)
    (9, 15.3)
    (10, 14.9)
    (11, 14.5)
    (12, 14.8)
    (13, 14.8)
    (14, 14.9)
    (15, 14.6)
    (16, 14.4)
    (17, 14.6)
    (18, 14.8)
	};
\addlegendentry{DTM}
\end{axis}
\end{tikzpicture}%
\caption{COH (mean) Over Time} \label{COH (mean) Over Time}
\end{subfigure}\hspace*{\fill}%
~%
\begin{subfigure}{0.25\textwidth}
\centering
\begin{tikzpicture}[scale=0.45,trim axis left, trim axis right]
\begin{axis}[
    xlabel={Years},
    ylabel={COH (median) of topics: COH  $\times$ $10^{-2}$},
    xmin=0, xmax=18,
    ymin=12.0, ymax=18.0,
    xtick={0,1,2,3,4,5,6,7,8,9,10,11,12,13,14,15,16,17,18},
    ytick={12.0,12.5,13.0,13.5,14.0, 14.5, 15.0, 15.5, 16.0, 16.5, 17.0,17.5,18.0},
    xticklabels={96,97,98,99,00,01,02,03,04,05,06,07,08,09,10,11,12,13,14},
    x tick label style={rotate=90,anchor=east},
    legend pos=north west,
    ymajorgrids=true,
    grid style=dashed,
]
\addplot[
    color=blue,
    mark=square,
    ]
    plot coordinates {
    (0, 16.0)
    (1, 16.1)
    (2, 16.2)
    (3, 15.6)
    (4, 16.2)
    (5, 15.2)
    (6, 15.5)
    (7, 15.4)
    (8, 15.6)
    (9, 15.0)
    (10, 15.2)
    (11, 15.5)
    (12, 15.2)
    (13, 15.2)
    (14, 15.4)
    (15, 15.5)
    (16,  15.7)
    (17, 15.5)
    (18, 15.5)
    };
\addlegendentry{RNN-RSM}
\addplot[
	color=gray,
	mark=diamond,
	]
	plot coordinates {
    (0, 13.8)
    (1, 14.0)
    (2, 13.6)
    (3, 13.2)
    (4, 13.6)
    (5, 13.3)
    (6, 15.8)
    (7, 15.8)
    (8, 15.7)
    (9, 14.2)
    (10, 14.8)
    (11, 13.7)
    (12, 13.7)
    (13, 13.9)
    (14, 13.6)
    (15, 13.1)
    (16, 12.7)
    (17, 12.8)
    (18, 14.3)
	};
\addlegendentry{DTM}
\end{axis}
\end{tikzpicture}%
\caption{COH (median) Over Time} \label{COH (median) Over Time}
\end{subfigure}\hspace*{\fill}%
~%
\begin{subfigure}{0.25\textwidth}
\centering
\begin{tikzpicture}[scale=0.45,trim axis left, trim axis right]
\begin{axis}[
    xlabel={Years},
    ylabel={Cosine Similarity},
    xmin=0, xmax=4,
    ymin=0.5, ymax=1.0,
    xtick={0,1,2,3,4},
    ytick={0.5,0.6,0.7,0.8,0.9,1.0},
    xticklabels={96-00, 00-05, 05-10, 10-14},
    x tick label style={rotate=45,anchor=east},
    legend pos=south east,
    ymajorgrids=true,
    grid style=dashed,
]
 
\addplot[
    color=blue,
    mark=square,
    ]
    plot coordinates {
    (0,0.74999999999999989)
    (1,0.69999999999999984)
    (2,0.8999999999999998)
    (3,0.99999999999999978)
    };
\addlegendentry{Sentiment Analysis}

\addplot[
	color=red,
	mark=*,
	]
	plot coordinates {
	(0,0.59999999999999987)
	(1,0.74999999999999989)
	(2,0.8999999999999998)
	(3,0.99999999999999978)
	};
\addlegendentry{Word Vector}

\addplot[
	color=Forest green,
	mark=x,
	]
	plot coordinates {
	(0,0.54999999999999993)
	(1,0.94999999999999984)
	(2,0.94999999999999984)
	(3,0.99999999999999978)
	};
\addlegendentry{Dependency Parsing}

\end{axis}
\end{tikzpicture}%
\caption{RNN-RSM Adj Topic Sim} \label{RNN-RSM Adjacent cosine similarity.}
\end{subfigure}\hspace*{\fill}%
~%
\begin{subfigure}{0.24\textwidth}
\centering
\begin{tikzpicture}[scale=0.45,trim axis left, trim axis right]
\begin{axis}[
    xlabel={Years},
    ylabel={Cosine Similarity},
    xmin=0, xmax=4,
    ymin=0.1, ymax=1.0,
    xtick={0,1,2,3,4},
    ytick={0.1,0.2,0.3,0.4,0.5,0.6,0.7,0.8,0.9,1.0},
    xticklabels={96-00, 00-05, 05-10, 10-14},
    x tick label style={rotate=45,anchor=east},
    legend pos=south east,
    ymajorgrids=true,
    grid style=dashed,
]
 \addplot[
    color=blue,
    mark=square,
    ]
    plot coordinates {
    (0,0.19999999999999996)
    (1,0.84999999999999987)
    (2,0.69999999999999984)
    (3,0.84999999999999987)
    };
\addlegendentry{Sentiment Analysis}

\addplot[
    color=red,
    mark=*,
    ]
    plot coordinates {
    (0,0.99999999999999978)
    (1,0.8999999999999998)
    (2,0.8999999999999998)
    (3,0.8999999999999998)
    };
\addlegendentry{Word Vector}

\addplot[
    color=Forest green,
    mark=x,
    ]
    plot coordinates {
    (0,0.8999999999999998)
    (1,0.79999999999999982)
    (2,0.74999999999999989)
    (3,0.8999999999999998)
    };
\addlegendentry{Dependency Parsing}
\end{axis}
\end{tikzpicture}%
\caption{DTM Adj Topic Sim} \label{DTM Adjacent cosine similarity.}
\end{subfigure}\hspace*{\fill}%

\medskip
\begin{subfigure}{0.25\textwidth}
\centering
\begin{tikzpicture}[scale=0.45,trim axis left, trim axis right]
    \begin{axis}[
        major x tick style = transparent,
        ybar=2*\pgflinewidth,
        bar width=8pt,
        ymajorgrids = true,
        ylabel = {Cosine Similarity},
        xlabel = {Starting Year},
        symbolic x coords={1996,2000,2005,2010},
        xtick = data,
        scaled y ticks = false,
        enlarge x limits=0.25,
        ymin=0.0,
        ymax=1.0,
        legend pos=north east
    ]
        \addplot[style={bblue,fill=bblue,mark=none}]
            coordinates {(1996, 0.59333333333333338) (2000,0) (2005, 0) (2010,0)};

        \addplot[style={rred,fill=rred,mark=none}]
             coordinates {(1996, 0.58333333333333337) (2000,0.60999999999999999) };

        \addplot[style={ggreen,fill=ggreen,mark=none}]
             coordinates {(1996, 0.18166666666666664) (2000,0.1900000000000005)  (2005, 0.18666666666666665) };

        \addplot[style={ppurple,fill=ppurple,mark=none}]
             coordinates {(1996, 0.2499999999999999)(2000,0.23333333333333339)(2005,0.25)(2010,0.23166666666666669)};

        \legend{2000,2005,2010,2014}
    \end{axis}
\end{tikzpicture}%
\caption{\small RSM} \label{RSM.}
\end{subfigure}\hspace*{\fill}%
~%
\begin{subfigure}{0.25\textwidth}
\centering
\begin{tikzpicture}[scale=0.45,trim axis left, trim axis right]
    \begin{axis}[
        major x tick style = transparent,
        ybar=2*\pgflinewidth,
        bar width=8pt,
        ymajorgrids = true,
        ylabel = {Cosine Similarity},
        xlabel = {Starting Year},
        symbolic x coords={1996,2000,2005,2010},
        xtick = data,
        scaled y ticks = false,
        enlarge x limits=0.25,
        ymin=0.0,
        ymax=1.0,
        legend pos=south east
    ]
        \addplot[style={bblue,fill=bblue,mark=none}]
            coordinates {(1996, 0.85906179648306613) (2000,0) (2005, 0) (2010,0)};

        \addplot[style={rred,fill=rred,mark=none}]
             coordinates {(1996, 0.8325525618335502)  (2000,0.85980319860048893)};

        \addplot[style={ggreen,fill=ggreen,mark=none}]
             coordinates {(1996, 0.78370485112268262) (2000,0.8018775831988203)  (2005, 0.84446196191669987) };

        \addplot[style={ppurple,fill=ppurple,mark=none}]
             coordinates {(1996, 0.78236204942496879)(2000,0.80829037686547611) (2005,0.79375389691564868)(2010,0.82981449286857778)};

        \legend{2000,2005,2010,2014}
    \end{axis}
\end{tikzpicture}%
\caption{\small LDA} \label{LDA.}
\end{subfigure}\hspace*{\fill}%
\begin{subfigure}{0.25\textwidth}
\centering
\begin{tikzpicture}[scale=0.45,trim axis left, trim axis right]
    \begin{axis}[
        major x tick style = transparent,
        ybar=2*\pgflinewidth,
        bar width=8pt,
        ymajorgrids = true,
        ylabel = {Cosine Similarity},
        xlabel = {Starting Year},
        symbolic x coords={1996,2000,2005,2010},
        xtick = data,
        scaled y ticks = false,
        enlarge x limits=0.25,
        ymin=0.0,
        ymax=1.0,
        legend pos=south east
    ]
        \addplot[style={bblue,fill=bblue,mark=none}]
            coordinates {(1996, 0.99318683826182952) (2000,0) (2005, 0) (2010,0)};

        \addplot[style={rred,fill=rred,mark=none}]
             coordinates {(1996, 0.96336249984691025)  (2000,0.9713411534141887)};

        \addplot[style={ggreen,fill=ggreen,mark=none}]
             coordinates {(1996, 0.93665586981575299) (2000,0.94336002989628687)  (2005, 0.97167417527325262) };

        \addplot[style={ppurple,fill=ppurple,mark=none}]
             coordinates {(1996, 0.91670526072451541)(2000,0.92401877060789939) (2005,0.94865496030916996)(2010,0.98159661469302917)};

        \legend{2000,2005,2010,2014}
    \end{axis}
\end{tikzpicture}%
\caption{\small DTM} \label{DTM.}
\end{subfigure}\hspace*{\fill}%
\begin{subfigure}{0.24\textwidth}
\centering
\begin{tikzpicture}[scale=0.45,trim axis left, trim axis right]
    \begin{axis}[
        major x tick style = transparent,
        ybar=2*\pgflinewidth,
        bar width=8pt,
        ymajorgrids = true,
        ylabel = {Cosine Similarity},
        xlabel = {Starting Year},
        symbolic x coords={1996,2000,2005,2010},
        xtick = data,
        scaled y ticks = false,
        enlarge x limits=0.25,
        ymin=0.0,
        ymax=1.0,
        legend pos=south east
    ]




      \addplot[style={bblue,fill=bblue,mark=none}]
            coordinates {(1996, 0.80247593083141422) (2000,0) (2005, 0) (2010,0)};

        \addplot[style={rred,fill=rred,mark=none}]
             coordinates {(1996, 0.75754122681320013)  (2000,0.85274685796822414)  (2005,0)};

        \addplot[style={ggreen,fill=ggreen,mark=none}]
             coordinates {(1996, 0.74109763735688983)(2000,0.82781743914808625) (2005, 0.95118820336782195)  };

        \addplot[style={ppurple,fill=ppurple,mark=none}]
             coordinates {(1996, 0.74254086683949427) (2000,0.82743092321994105) (2005,0.95108360558704907)(2010,0.9805636356589056)};

        \legend{2000,2005,2010,2014}
    \end{axis}
\end{tikzpicture}%
\caption{\small RNN-RSM} \label{RNN-RSM.}
\end{subfigure}\hspace*{\fill}%
\caption{{(a, b, c)}: Topic popularity by LDA, RSM, DTM and RNN-RSM over time 
{(d)}: Perplexity on the unobserved document collections over time 
{(e, f)}: Mean and Median Topic Coherence  
{(g, h)}: Topic Evolution {(i,j,k,l)}: Topic focus change over time. Adj- Adjacent; Sim- Similarity} \label{fig:topic popularity}
\end{figure*}
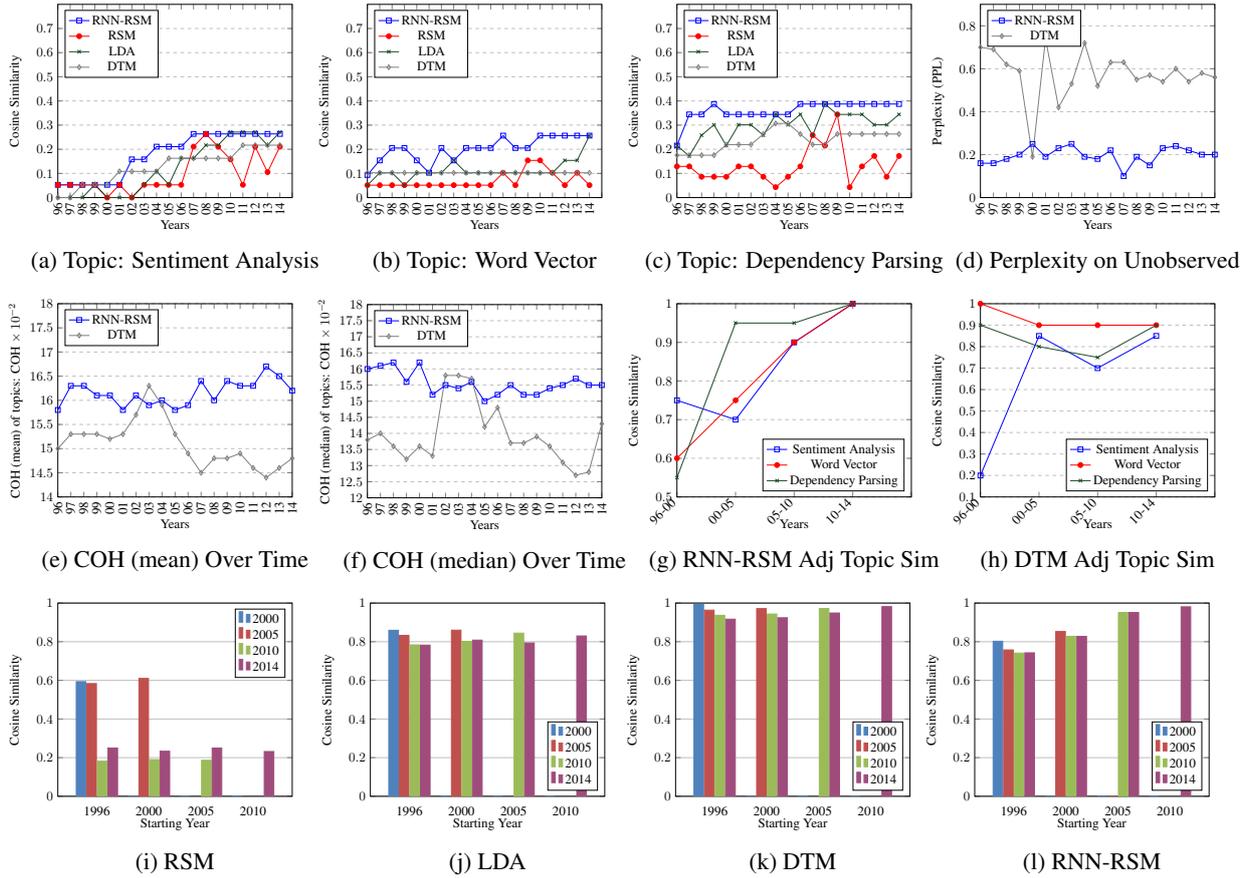

\section{Evaluation}
\subsection{Dataset and Experimental Setup}
We use the processed dataset~\cite{Gol:82}, consisting of EMNLP and ACL conference papers from the year 1996 through 2014 (Table~\ref{paperscount}).  
We combine papers for each year 
from the two venues to prepare the document collections over time.  
We use  ExpandRank~\cite{Wan:82} to extract top 100 keyphrases for each paper, including unigrams and bigrams.
We split the bigrams to unigrams to create a dictionary of all unigrams and bigrams. 
The dictionary size ($K$) and word count are 3390 and 5.19 M, respectively. 

We evaluate RNN-RSM against static (RSM, LDA) and dynamic (DTM) topics models for topic and keyword evolution in NLP research over time. 
Individual 19 different RSM and LDA models are trained 
for each year, while DTM\footnote{https://radimrehurek.com/gensim/models/dtmmodel.html} and RNN-RSM are trained over the years with 19 time steps, 
where paper collections for a year is input at each time step.
RNN-RSM is initialized with RSM ({\small ${\bf W_{vh}}$, ${\bf b_{v}}$, ${\bf b_{h}}$}) 
trained for the year 2014. 
\begin{table}[t]
\centering
\def\arraystretch{1.18}
\resizebox{0.4\textwidth}{!}{
\begin{tabular}{c|c|c}\hline
 {\bf Parameter}  & {\bf Value(s)} & {\bf Optimal}\\
\hline
{\it epochs} &  1000 & 1000\\  
{\it CD  iterations} &   15 & 15\\
{\it learning rate} & 0.1, 0.03, 0.001 & 0.001\\
{\it hidden size} & 20, 30, 50 & 30 \\ \hline
\end{tabular}}
\caption{Hyperparameters for RNN-RSM model}
\label{hyperparamters}
\end{table}

We use perplexity to 
choose the number of topics (=30). 
See Table \ref{hyperparamters} for hyperparameters. 
\begin{table}[t]
\centering
\def\arraystretch{1.25}
\resizebox{0.48\textwidth}{!}{
\begin{tabular}{c|c|c||c|c||c}
\hline
\multirow{2}{*}{\bf model}     & \multicolumn{5}{c}{\bf metric}\\ \cline{2-6}
                           & {\it SumPPL}                               & {\it Err}                              & {\it mean-COH}         & {\it median-COH}     & {\it TTD}   \\  \hline 
{\it DTM}           & 10.9                &  8.10                        &   0.1514                  &   0.1379                            &         0.084       \\
{\it RNN-RSM}   & {\bf 3.8}      & {\bf 7.58}                &  {\bf 0.1620}           &  {\bf 0.1552}                   &       \underline{0.268}         \\     \hline
\end{tabular}}
\caption{State-of-the-art Comparison: Generalization ({\it PPL} and {\it Err}), Topic Interpretation ({\it COH}) and Evolution ({\it TTD}) in DTM and RNN-RSM models}
\label{generalization}
\end{table}
\subsection{Generalization in Dynamic Topic Models}
{\bf Perplexity:} We compute the perplexity on unobserved documents ($\widehat{{\bf V}}^{(t)}$) at each time step as \vspace*{-0.25cm}
\begin{equation*}
\mbox{PPL}(\widehat{{\bf V}}^{(t)}, t) = \exp\Big(- \frac{1}{N^{(t)}} \frac{\sum_{n=1}^{N^{(t)}} \log P({\bf V}_{n}^{(t)}) }{\sum_{n=1}^{N^{(t)}} D_{n}^{(t)} }\Big)
\end{equation*}
where $t$ is the time step. $N^{(t)}$ is the number of documents in a collection ($\widehat{\bf V}^{(t)}$) at time $t$. 
Better models have lower perplexity values, suggesting less uncertainties about the documents.
For held-out documents, we take 10 documents from each time step i.e. total 190 documents and compute perplexity for 30 topics. 
Fig \ref{Perplexity over time unobserved} shows the comparison of perplexity values for unobserved documents from DTM and RNN-RSM at each time step. 
The {\it SumPPL} (Table \ref{generalization}) is the sum of PPL values for the held-out sets of each time step.\\
\begin{table*}[t]
\centering
\def\arraystretch{1.1}
\resizebox{\textwidth}{!}{
\begin{tabular}{c|c|c}
\hline
{\bf Drift}  & {\bf Model (year)} & {\bf Topic Terms} \\ \hline
\multirow{4}{*}{0.20}    &  \multirow{2}{*}{{\bf DTM} (1996)}                 
&  document, retrieval, query, documents, information, search, information retrieval, queries, terms,  \\
& & words, system, results, performance, method, approach \\ \cline{2-3}
& \multirow{2}{*}{{\bf DTM} (2014)}    
&  document, query, search, documents, queries, information, retrieval, method, results, \\
& &  information retrieval, research, terms, other, approach,  knowledge \\  \hline 
\multirow{4}{*}{0.53}    &  \multirow{2}{*}{{\bf DTM} (1996)}                 
&  semantic, lexical, structure, syntactic, argument, frame, example, lexicon, information, approach, \\ 
& & source, function, figure, verbs, semantic representation   \\ \cline{2-3}
& \multirow{2}{*}{{\bf DTM} (2014)}   
& semantic, argument, frame, sentence, syntactic, semantic parsing, structure, semantic role,  \\ 
& & example, role labeling, language, learning, logical form, system, lexicon  \\  \hline \hline
\multirow{4}{*}{0.20}    &  \multirow{2}{*}{{\bf RNN-RSM} (1996)}         
&    reordering, statistical machine, translation model, translations, arabic,  word align, translation probability, word alignment, \\ 
& & translation system, source word, ibm model, source sentence, english translation, target language, word segmentation \\ \cline{2-3}
& \multirow{2}{*}{{\bf RNN-RSM} (2014)}    &  reordering, statistical machine, translation model, translations, arabic,  word align, translation probability, word alignment, \\
& & translation system, source word, reordering model,  bleu score, smt system, english translation, target language  \\ \hline
\multirow{4}{*}{0.53}    &  \multirow{2}{*}{{\bf RNN-RSM} (1996)}                
 &     input, inference, semantic representation, distributional models, logical forms, space model,  clustering algorithm, space models, \\
& &  similar word, frequent word, meaning representation, lexical acquisition, new algorithm, same context, multiple words\\ \cline{2-3}
& \multirow{2}{*}{{\bf RNN-RSM} (2014)}    
&   input, inference, word vector, word vectors, vector representation, semantic representation, distributional models, semantic space, \\
& &  space model, semantic parser, vector representations, neural language, logical forms, cosine similarity, clustering algorithm\\ \hline 
\end{tabular}}
\caption{Topics (top 15 words) with the highest and lowest drifts (cosine) observed in DTM and RNN-RSM}
\label{drifts for topics}
\end{table*}
{\bf Document Time Stamp Prediction:}
To further assess the dynamic topics models, we split the document collections at each time step into 80-20\% train-test, resulting in $1067$ held-out documents. 
We predict the time stamp  (dating) of a document 
by finding the most likely (with the lowest perplexity) location over the time line. 
See the {\it mean absolute error} ({\it Err}) in year for the held-out in Table \ref{generalization}. 
Note, we do not use the time stamp as observables during training. 
\subsection{TSD, TED: Topic Evolution over Time}
{\bf Topic Detection:} To extract topics from each RSM, we compute posterior $P(\widehat{\bf V}^{(t)}|h_{j}=1)$ 
by activating a hidden unit and deactivating the rest in a hidden layer. 
We extract the top 20 terms for every 30 topic set from 1996-2014, 
resulting in $|Q|_{max} = 19 \times 30 \times 20$ possible topic terms. 

{\bf Topic Popularity:} To determine topic {\it popularity}, we selected three popular topics ({\it Sentiment Analysis}, {\it Word Vector} and {\it Dependency Parsing}) in NLP research and create a set\footnote{topic-terms to be released with code} of key-terms (including unigrams and bigrams) for each topic. 
We compute cosine similarity of the key-terms defined for each selected topic and  topics discovered by the topic models over the years. 
We consider the discovered topic that is the most similar to the key-terms in the target topic and plot the similarity values in Figure~\ref{Standard cosine similarity Sentiment Analysis.}, \ref{Standard cosine similarity Word Vector.} and \ref{Standard cosine similarity Word Vector.}. 
Observe that RNN-RSM shows better topic evolution for the three emerging topics. 
LDA and RSM show topical locality in Figure~\ref{Standard cosine similarity Dependency Parsing.} attributed to no correlation in topic dynamics over time,  
while in Figure~\ref{Standard cosine similarity Word Vector.}, DTM does not capture evolution of topic {\it Word Vector}.

{\bf Topic Drift (Focus Change):} To compute the topic {\it focus} change over the years, we first split the time period 1996-2014 into five parts:\{1996, 2000, 2005, 2010, 2014\}.
The cosine similarity scores are computed between the topic sets discovered in a particular year and the years preceding it in the above set, 
for example  the similarity scores between the topic-terms in (1996, 2000), (1996, 2005), (1996, 2010) and (1996, 2014), respectively.  
Figure~\ref{RSM.}, \ref{LDA.}, \ref{DTM.} and \ref{RNN-RSM.} demonstrate that RNN-RSM shows higher {\it convergence} in topic focus over the years, compared to LDA and RSM.
In RNN-RSM, the topic similarity is gradually increased over time, however not in DTM. 
The higher similarities in the topic sets indicate that new/existing topics and words do not appear/disappear over time.  

We compute topic-term drift ($TTD$) to show the changing topics from initial to final year, as
\begin{equation*}
TTD = 1.0 -  cosineSimilarity({\bf Q}^{(t)}, {\bf Q}^{(t')})
\end{equation*}
where {\bf Q} is the set of all topic-terms for time step $t$. 
Table \ref{generalization} shows that $TTD$ (where $t$=$1996$ and $t'$=$2014$) are $0.268$ and $0.084$ for RNN-RSM and DTM, respectively. 
It suggests that the higher number of new topic-terms evolved in RNN-RSM, compared to DTM.
Qualitatively, the Table \ref{drifts for topics} shows the topics observed with the highest and lowest cosine drifts in DTM and RNN-RSM.

In Figure~\ref{RNN-RSM Adjacent cosine similarity.} and \ref{DTM Adjacent cosine similarity.}, we also illustrate the temporal {\it evolution} (drift) in the selected  topics   
by computing cosine similarity on their adjacent topic vectors over time. 
The topic vectors are selected similarly as in computing topic popularity. 
We observe better TED in RNN-RSM than DTM for the three emerging topics in NLP research. 
For instance, for the selected topic {\it Word Vector},  the red line in DTM (Fig~\ref{DTM Adjacent cosine similarity.}) 
shows no drift  (for x-axis 00-05, 05-10 and 10-14), 
suggesting the topic-terms in the adjacent years  are similar and does not evolve.

\subsection{Topic Interpretability}
Beyond perplexities, 
we also compute topic coherence \cite{Chang:82, Newman:82, Das:82} to determine the meaningful topics captured. 
We use the coherence measure proposed by \citeauthor{Aletras:82} \shortcite{Aletras:82} that retrieves co-occurrence counts for the set of topic words 
using Wikipedia as a reference corpus to identify context features (window=5) for each topic word.  
Relatedness between topic words and context features is measured using normalized pointwise mutual information (NPMI), 
resulting in a single vector for every topic word. The coherence ({\it COH}) score is computed as the arithmetic mean of the cosine similarities between all word pairs.
Higher scores imply more coherent topics. 
We use Palmetto\footnote{\url{github.com/earthquakesan/palmetto-py}} library to estimate coherence.\\
{\bf Quantitative:} 
We compute mean and median coherence scores for each time step using the corresponding topics, as shown in Fig \ref{COH (mean) Over Time} and \ref{COH (median) Over Time}. 
Table \ref{generalization} shows {\it mean-COH} and {\it median-COH} scores, computed by mean and median of scores from  Fig~\ref{COH (mean) Over Time} and \ref{COH (median) Over Time}, respectively.
Observe that RNN-RSM captures topics with higher coherence.   \\
{\bf Qualitative:}
Table \ref{PMI for topics} shows topics (top-10 words) with the highest and lowest coherence scores. 

\subsection{TTC: Trending Keywords over time}\label{TTC}
We demonstrate the capability of RNN-RSM to capture 
word evolution (usage) in topics over time.  
We define: {\it keyword-trend} and SPAN.
The {\it keyword-trend} is  the appearance/disappearance of the keyword in topic-terms detected over time, while 
 SPAN is the length of the longest sequence of the keyword appearance in its keyword trend. 

Let $\widehat{\bf Q}_{model} = \{{\bf Q}_{model}^{(t)}\}_{t=1}^{T}$ be a set of 
sets\footnote{a set by ${\bf bold}$ and set of sets by $\widehat{\bf bold}$}  
of topic-terms discovered by the $model$ (LDA, RSM, DTM and RNN-RSM) over different time steps. 
Let {${\bf Q}^{(t)} \in  \widehat{\bf Q}_{model}$} be the topic-terms at time step $t$. 
The keyword-trend for a keyword $k$ is a time-ordered sequence of 0s and 1s, as\vspace*{0.15cm}

\begin{table}[t]
\centering
\def\arraystretch{1.1}
\resizebox{0.48\textwidth}{!}{
\begin{tabular}{c|c||c|c}
\hline
{\bf DTM} (2001) & {\bf  RNN-RSM} (2001) & {\bf DTM} (2012)  & {\bf RNN-RSM} (1997) \\ \hline
semantic            &       words         &      discourse     &  parse \\
frame            &       models          &      relation      & cluster\\
argument            &       grammar          &   relations       &    clustering\\
syntactic            &       trees         &     structure      &    results \\
structure            &       dependency parsing          &   sentence        &    query \\
lexical            &       parsers          &        class     &  pos tag  \\
example            &      dependency trees           &   lexical       &    queries \\
information            &    parsing            &    argument         &  retrieval \\
annotation            &      parse trees            &  corpus        &    coreference \\
lexicon            &      dependency parse           &     other       &  logical form \\ \hline
\multicolumn{1}{l|}{{\bf COH:}  \, 0.268} & \underline{0.284} & 0.064 & \underline{0.071} \\ \hline
\end{tabular}}
\caption{Topics with the highest and lowest coherence}
\label{PMI for topics}
\end{table}

\texttt{\noindent \noindent \phantom{x}\hspace{7.5ex}}{$\mbox{trend}_{k}(\widehat{\bf Q}) = [\mbox{find} (k,  {\bf Q}^{(t)})]_{t=1}^{T} $
\begin{align}
\begin{split}\label{eq:find}
\mbox{where}; \ \ \mbox{find} (k, {\bf Q}^{(t)}) =\begin{cases}
1  \quad \text{if $k \in {\bf Q}^{(t)}$}&\\ 
0 \quad \text{otherwise}&
\end{cases}
\end{split}
\end{align}}

And the SPAN ($S_k$) for the $k$th keyword is-\vspace*{0.15cm}

{${S}_{k}(\widehat{\bf Q}) = \mbox{length}\big(\mbox{longestOnesSeq}(\mbox{trend}_{k}(\widehat{\bf Q})\big)$}\vspace*{0.15cm}

We compute keyword-trend and SPAN for each term from the set of some popular terms. 
We define average-SPAN for all the topic-terms appearing in the topics discovered over the years,
\begin{align*}
\begin{split}
\mbox{{\small avg-SPAN}}&(\widehat{\bf Q}) = \frac{1}{||\widehat{\bf Q}||}\sum_{\{k | {\bf Q}^{(t)} \in \widehat{\bf Q} \land k \in {\bf Q}^{(t)}\}}\frac{{{\small S}}_{k}(\widehat{\bf Q})}{\hat{v}^{k}}
\end{split}\\
\begin{split}
&= \frac{1}{||\widehat{\bf Q}||}\sum_{\{k | {\bf Q}^{(t)} \in \widehat{\bf Q} \land k \in {\bf Q}^{(t)} \}} {{\small S}}_{k}^{dict} (\widehat{\bf Q})
\end{split}
\end{align*}
where {$||\widehat{\bf Q}|| = | \{k | {\bf Q}^{(t)} \in \widehat{\bf Q} \land k \in {\bf Q}^{(t)} \} |$} 
is the count of unique topic-terms and 
{${\hat{v^{k}}}$ = $ \sum_{t=1}^{T}\sum_{j=1}^{D_{t}}v_{j, t}^{k}$} 
denotes the count of  $k^{th}$ keyword. 

\begin{figure}[t]
{
  \centering
   \includegraphics[scale = 0.78]{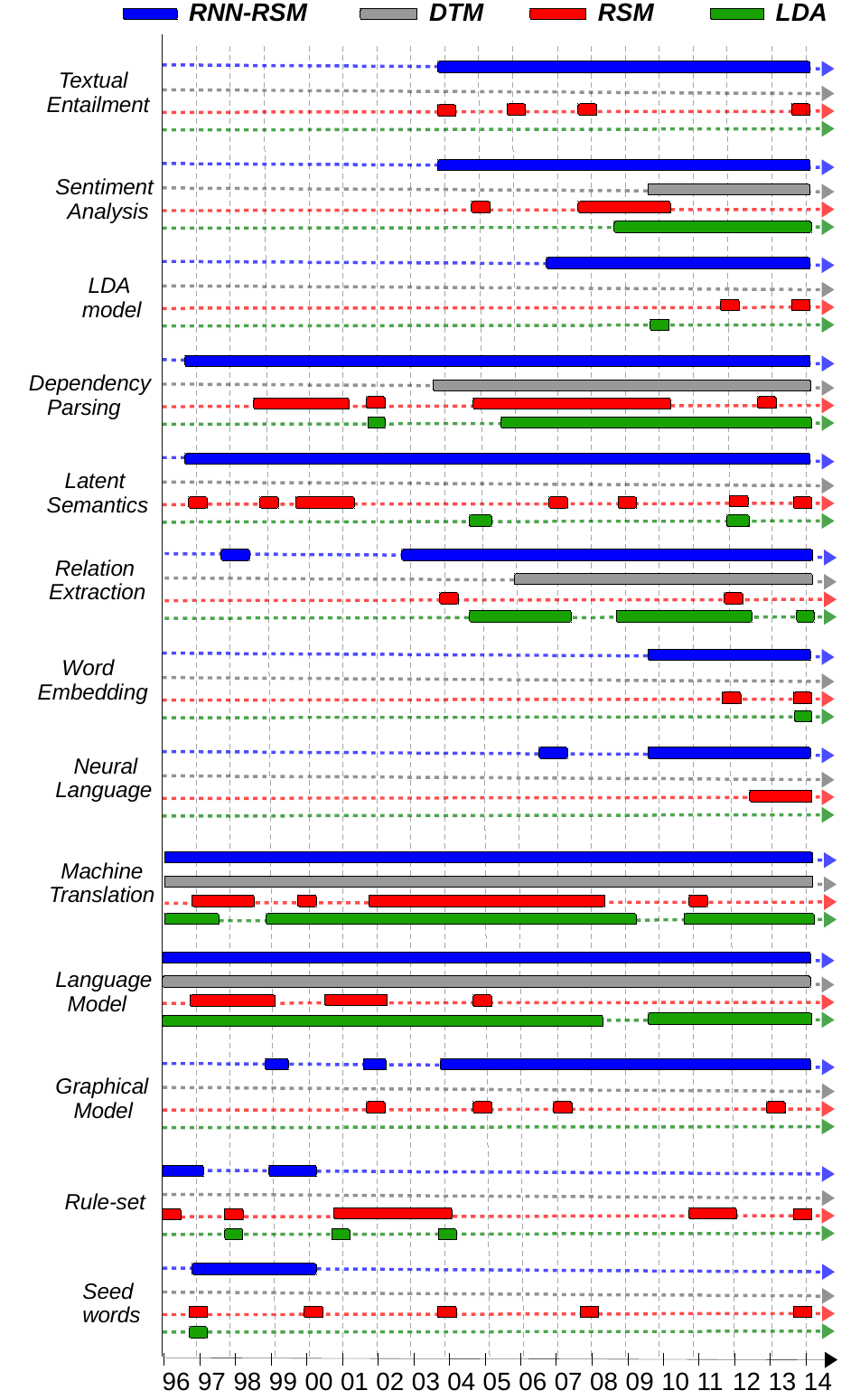}
    \caption[Keyword Trend Detection by RNN-RSM, DTM, RSM and LDA models.]
    {Keyword-trend by RNN-RSM, DTM, RSM, LDA. 
     Bar: Keyword presence in topics for the year}
    \label{fig:Keyword Trend Detection: RSM and RNN-RSM.}
}
\end{figure}
\begin{figure}[t]
{
  \centering
   \includegraphics[scale = 0.35]{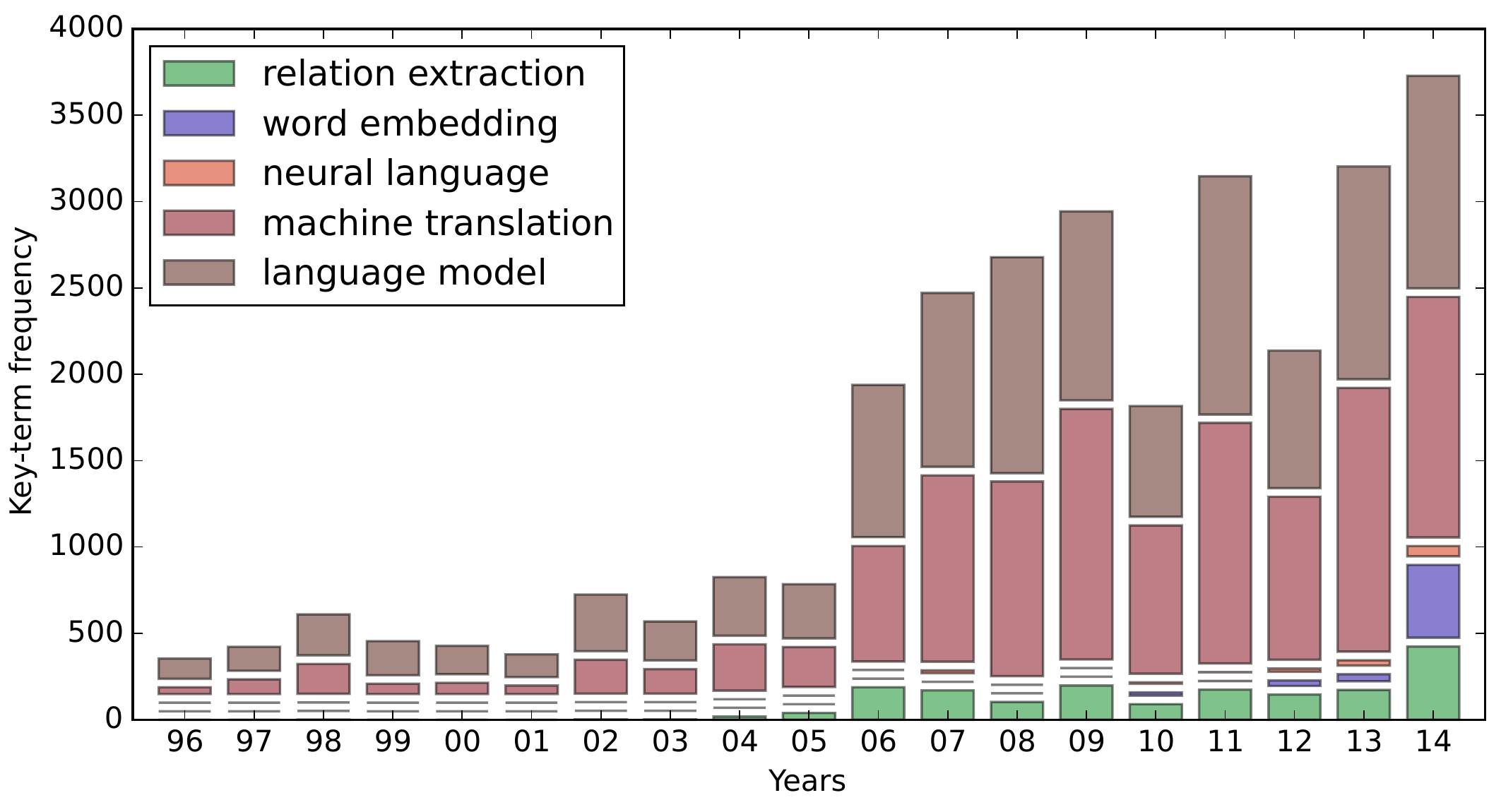}
    \caption{Key-term frequency in the input over years}
    \label{fig:Inputwordfreq}
}
\end{figure} 

In Figure \ref{fig:Keyword Trend Detection: RSM and RNN-RSM.}, the keyword-trends indicate emergence (appearance/disappearance) 
of the selected popular terms in topics discovered in  ACL and EMNLP papers over time. 
Observe that RNN-RSM captures longer SPANs for popular keywords and better word usage in NLP research.   
For example: {\it Word Embedding} is one of the top keywords, appeared locally (Figure \ref{fig:Inputwordfreq}) in the recent years. 
RNN-RSM detects it in the topics from 2010 to 2014, however DTM does not. 
Similarly, for {\it Neural Language}. However, {\it Machine Translation} and {\it Language Model} are globally appeared in the input document collections over time  
and captured in the topics by RNN-RSM and DTM.  
We also show keywords ({\it Rule-set} and {\it Seed Words}) 
that disappeared in topics over time.

\begin{table}[t]
\centering
\def\arraystretch{1.2}
\resizebox{0.48\textwidth}{!}{
\begin{tabular}{l|c|cc|cc|cc|cc}
\hline
\multicolumn{1}{c|}{\multirow{2}{*}{\bf Term}} & \multirow{2}{*}{\bf $\hat{v^{k}}$}  & \multicolumn{2}{c|}{\bf LDA} & \multicolumn{2}{c|}{\bf RSM} & \multicolumn{2}{c|}{\bf DTM} & \multicolumn{2}{c}{\bf RNN-RSM} \\
\multicolumn{1}{c|}{}                      &                    & $S_{k}$   &$S_{k}^{dict}$   & $S_{k}$   &$S_{k}^{dict}$ & $S_{k}$   &$S_{k}^{dict}$ & $S_{k}$ &$S_{k}^{dict}$    \\
\hline
  Textual entailment                                        &  918                  & 0      &         .000           &    1     &               .001               &    0     &               .000               &    {\bf 11}       &          .011     \\
  Sentiment analysis                                        &  1543               &  6     &         .004             &   3      &    .002                         &    5     &               .0032               &     {\bf  11}      &           0.007      \\
  Lda model                                                      &    392               &  1   &            .003         &   1    &         .002                     &    0     &               .000               &     {\bf  8}      &            .020   \\
  Dependency parsing                                     &    3409            &   9    &          .003              &  5       &              .001                &    11     &               .0032               &    {\bf 18}      &         .005       \\
  Latent semantic                                             &    974               &  1     &          .001           &    2     &          .002                    &    0    &               .000               &     {\bf 18}     &          .018      \\
  Relation extraction                                        &  1734              &  4     &           .002             &    1     &        .001                      &    9     &               .0052               &      {\bf 12}     &          .007        \\
  Word embedding                                           &    534              & 1      &          .002            &   1     &                .002              &    0    &               .000               &      {\bf  5}    &                 .009 \\
  Neural language                                           &   121             &   0    &            .000           &     3    &                  .025            &    0    &               .000               &      {\bf 5}    &               .041    \\
  Machine translation                                     &     11741           &  11     &      .001                 &   7      &               .001               &    {\bf 19}     &               .0016               &     {\bf 19}      &        .002         \\
  Language model                                           &    11768             &   13    &     .001                  &  3       &               .000              &    {\bf 19}     &               .0016               &    {\bf 19}       &         .002       \\  
  Graphical model                                           &       680          &   0    &     .000                  &  1       &               .001              &    0     &               .000               &    {\bf  11}       &         .016       \\  
  Rule set                                                      &     589            &   1   &     .0017    &   {\bf   4}   &      .0068          &     0   &            .000                 &     2     &     .0034 \\
  Seed words                                                &  396            &   1   &     .0025    &     1   &      .0025          &     0   &            .000                 &     {\bf 4}     &     .0101      \\
\hline
\hline 
\multicolumn{1}{c}{$\mbox{avg-SPAN}(\widehat{\bf Q})$}                                         &                   &       &      .002               &         &     .007                    &         &     .003                    &          &         {\bf .018}      \\
\hline
\multicolumn{1}{c}{$||\widehat{\bf Q}_{model}||$ }                                       &  &  \multicolumn{2}{|c}{926}       &      \multicolumn{2}{|c}{2274}                   &         \multicolumn{2}{|c}{ \underline{335}}                    &           \multicolumn{2}{|c}{ \underline{731}}      
\end{tabular}
}
\caption{\label{Keyword span comparison: RSM and RNN-RSM.}SPAN ($S_k$) for selected terms, avg-SPAN and set $||\widehat{\bf Q}||$ by LDA, RSM, DTM and RNN-RSM}
\end{table}

 \begin{figure}[t]
       \centering
        \includegraphics[scale=0.69]{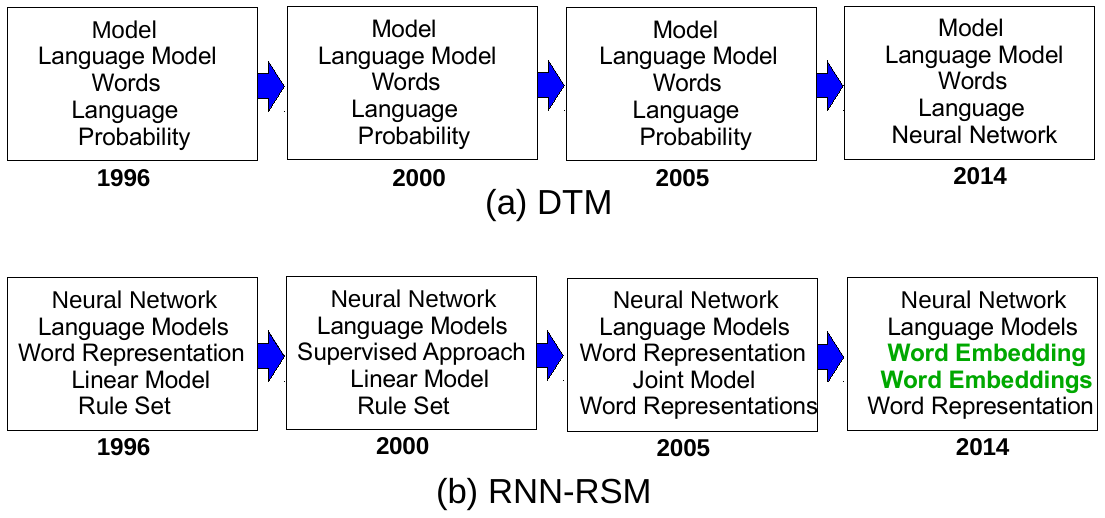}
        \captionof{figure}{Word usage for emerging topic {\it Word Vector} over time, captured by DTM and RNN-RSM}
        \label{wordsovertimeresults}
\end{figure}

Higher SPAN suggests that the model is capable in capturing trending keywords. 
Table \ref{Keyword span comparison: RSM and RNN-RSM.} shows corresponding comparison of SPANs for the 
13 selected keywords. The SPAN ${S}_{k}$ for each keyword is  computed  from Figure \ref{fig:Keyword Trend Detection: RSM and RNN-RSM.}. 
Observe that {$||\widehat{\bf Q}||_{DTM} < ||\widehat{\bf Q}||_{RNN-RSM}$} suggests new topics and words emerged over time in RNN-RSM, 
while higher SPAN values in RNN-RSM suggest better trends. 
Figure \ref{wordsovertimeresults} shows how the word usage, captured by DTM and RNN-RSM for 
the topic {\it Word Vector}, changes over 19 years in NLP research. RNN-RSM captures 
popular terms {\it Word Embedding} and {\it Word Representation} emerged in it.

\section{Discussion: RNN-RSM vs DTM}\label{RNN-RSm vs DTM}
{\bf Architecture:} RNN-RSM treats document's stream as high dimensional sequences over time and models the complex 
conditional probability distribution i.e. {\it heteroscedasticity} in document collections and topics over time by a temporal stack of RSMs (undirected graphical model), 
conditioned on time-feedback connections using RNN \cite{rum:82}.  
It has two hidden layers: {\bf h} (stochastic binary) to 
capture topical information, 
while {\bf u} (deterministic) to convey temporal information via BPTT 
that 
models the topic dependence at a time step $t$ on {\it all} the previous steps $\tau< t$. 
In contrast, DTM is built upon LDA (directed model), 
where Dirichlet distribution on words is not amenable to sequential modeling, 
therefore its natural parameters (topic and topic proportion distributions) for each topic are chained, 
instead of latent topics 
that results in intractable inference in topic detection and chaining.

{\bf Topic Dynamics:} 
The introduction of explicit connection in latent topics 
in RNN-RSM allow new topics and words for the underlying topics to appear or disappear over time 
by the dynamics of topic correlations. 
As discussed, the distinction of {\bf h} and {\bf u} permits the latent topic ${\bf h}^{(t)}$ 
to capture new topics, that may not be captured by ${\bf h}^{(t-1)}$.

DTM assumes a fixed number of global topics and models their distribution over time. 
However, there is no such assumption in RNN-RSM. 
We fixed the topic count in RNN-RSM at each time step, since ${\bf W_{vh}}$ is fixed over time 
and RSM biases turn off/on terms in each topic.
However, this is fundamentally different for DTM. 
E.g. a unique label be assigned to each of the $30$ topics at any time steps $t$ and $t'$. 
DTM follows the sets of topic labels: 
{$\{TopicLabels^{(t)}\}_{k=1}^{30} = \{TopicLabels^{(t')}\}_{k=1}^{30}$},  
due to eq (1) in \citeauthor{blei:83} \shortcite{blei:83} (discussed in section 5) that limits DTM to 
capture new (or local) topics or words appeared over time. 
It corresponds to the keyword-trends (section \ref{TTC}).

{\bf Optimization:} The RNN-RSM is based on Gibbs sampling and BPTT for inference while DTM employs complex variational methods, since applying Gibbs sampling is difficult due to the nonconjugacy of the Gaussian and multinomial distributions.  
Thus, easier learning in RNN-RSM.

For all models, approximations are solely used to compute the likelihood, either using variational approaches or contrastive divergence; 
perplexity was then computed based on the approximated likelihood. 
More specifically, we use variational approximations to compute the likelihood for DTM \cite{blei:83}. 
For RSM and RNN-RSM, the respective likelihoods are approximated using the standard Contrastive Divergence (CD). 
While there are substantial differences between variational approaches and CD, and 
thus in the manner the likelihood for different models is estimated - 
both approximations work well for the respective family of models in terms of approximating the true likelihood. 
Consequently, perplexities computed based on these approximated likelihoods are indeed comparable.

\section{Conclusion and Future Work}
We have proposed a neural temporal topic model which we name as RNN-RSM,  
based on probabilistic undirected graphical topic model RSM with time-feedback connections via deterministic RNN,
to capture temporal relationships in historical documents. 
The model is the first of its kind that learns topic dynamics in collections of different-sized documents over time, within the generative and neural network framework. 
The experimental results have demonstrated that RNN-RSM shows better generalization (perplexity and time stamp prediction), topic interpretation (coherence) and evolution (popularity and drift) 
in scientific articles over time.
We also introduced SPAN to illustrate topic characterization. 

In future work, we forsee to investigate learning dynamics in variable number of topics over time. 
It would also be an interesting direction to investigate the effect of the skewness in the distribution of papers over all years. 
Further, we see a potential application of the proposed model in learning the time-aware i.e. dynamic word embeddings \cite{Jean:85,Basile:85,Bamler:85,Rudolph:85,Yao:85} 
in order to capture language evolution over time, instead of document topics.  


\section*{Acknowledgments}
We thank Sujatha Das Gollapalli for providing us with the data sets used in the experiments. 
We express appreciation for our colleagues  Florian Buettner,  Mark Buckley,  Stefan Langer, Ulli Waltinger and Usama Yaseen, and 
anonymous reviewers for their in-depth review comments. 
This research was supported by Bundeswirtschaftsministerium ({\tt bmwi.de}), grant 01MD15010A (Smart Data Web) 
at Siemens AG- CT Machine Intelligence, Munich Germany. 

\bibliography{naaclhlt2018}

\begin{thebibliography}{}
\expandafter\ifx\csname natexlab\endcsname\relax\def\natexlab#1{#1}\fi

\bibitem[{Aitchison(2001)}]{Jean:85}
Jean Aitchison. 2001.
\newblock {\em Language change: progress or decay?\/}.
\newblock Cambridge University Press.

\bibitem[{Aletras and Stevenson(2013)}]{Aletras:82}
Nikolaos Aletras and Mark Stevenson. 2013.
\newblock Evaluating topic coherence using distributional semantics.
\newblock In {\em Proceedings of the 10th International Conference on
  Computational Semantics (IWCS)\/}. Potsdam, Germany, pages 13--22.

\bibitem[{Allan(2002)}]{Allan:83}
James Allan. 2002.
\newblock Introduction to topic detection and tracking.
\newblock In {\em Topic detection and tracking\/}, Springer, pages 1--16.

\bibitem[{Allan et~al.(1998)Allan, Carbonell, Doddington, Yamron, and
  Yang}]{Allan:82}
James Allan, Jaime~G Carbonell, George Doddington, Jonathan Yamron, and Yiming
  Yang. 1998.
\newblock Topic detection and tracking pilot study final report.
\newblock In {\em Proceedings of the DARPA Broadcast News Transcription and
  Understanding Workshop\/}. Virginia, US, pages 194--218.

\bibitem[{Bamler and Mandt(2017)}]{Bamler:85}
Robert Bamler and Stephan Mandt. 2017.
\newblock Dynamic word embeddings.
\newblock In {\em Proceedings of the 34th International Conference on Machine
  Learning\/}. Sydney, Australia, pages 380--389.

\bibitem[{Basile et~al.(2014)Basile, Caputo, and Semeraro}]{Basile:85}
Pierpaolo Basile, Annalina Caputo, and Giovanni Semeraro. 2014.
\newblock Analysing word meaning over time by exploiting temporal random
  indexing.
\newblock In {\em Proceedings of the 1st Italian Conference on Computational
  Linguistics (CLiC-it)\/}. Pisa University Press, Pisa, Italy.

\bibitem[{Blei and Lafferty(2006)}]{blei:83}
David~M. Blei and John~D. Lafferty. 2006.
\newblock Dynamic topic models.
\newblock In {\em Proceedings of the 23rd International Conference on Machine
  Learning\/}. Association for Computing Machinery, Pittsburgh, Pennsylvania
  USA, pages 113--120.

\bibitem[{Blei et~al.(2003)Blei, Ng, and Jordan}]{Blei:81}
David~M. Blei, Andrew~Y. Ng, and Michael~I. Jordan. 2003.
\newblock Latent dirichlet allocation.
\newblock {\em Proceedings of Machine Learning Research\/} 3(Jan):993--1022.

\bibitem[{Boulanger-Lewandowski et~al.(2012)Boulanger-Lewandowski, Bengio, and
  Vincent}]{nic:82}
Nicolas Boulanger-Lewandowski, Yoshua Bengio, and Pascal Vincent. 2012.
\newblock Modeling temporal dependencies in high-dimensional sequences:
  Application to polyphonic music generation and transcription.
\newblock In {\em Proceedings of the 29th International Conference on Machine
  Learning\/}. Edinburgh, Scotland UK.

\bibitem[{Chang et~al.(2009)Chang, Gerrish, Wang, Boyd-Graber, and
  Blei}]{Chang:82}
Jonathan Chang, Sean Gerrish, Chong Wang, Jordan~L. Boyd-Graber, and David~M.
  Blei. 2009.
\newblock Reading tea leaves: How humans interpret topic models.
\newblock In {\em Advances in Neural Information Processing Systems\/}. Curran
  Associates, Inc., Vancouver, Canada, pages 288--296.

\bibitem[{Das et~al.(2015)Das, Zaheer, and Dyer}]{Das:82}
Rajarshi Das, Manzil Zaheer, and Chris Dyer. 2015.
\newblock Gaussian lda for topic models with word embeddings.
\newblock In {\em Proceedings of the 53rd Annual Meeting of the Association for
  Computational Linguistics and the 7th International Joint Conference on
  Natural Language Processing\/}. Association for Computational Linguistics,
  Beijing, China, volume~1, pages 795--804.

\bibitem[{Gehler et~al.(2006)Gehler, Holub, and Welling}]{Welling:85}
Peter~V. Gehler, Alex~D. Holub, and Max Welling. 2006.
\newblock The rate adapting poisson model for information retrieval and object
  recognition.
\newblock In {\em Proceedings of the 23rd International Conference on Machine
  Learning\/}. Association for Computing Machinery, Pittsburgh, Pennsylvania
  USA, pages 337--344.

\bibitem[{Gollapalli and Li(2015)}]{Gol:82}
Sujatha~Das Gollapalli and Xiaoli Li. 2015.
\newblock Emnlp versus acl: Analyzing nlp research over time.
\newblock In {\em Proceedings of the Conference on Empirical Methods in Natural
  Language Processing\/}. Association for Computational Linguistics, Lisbon,
  Portugal, pages 2002--2006.

\bibitem[{Gupta et~al.(2015{\natexlab{a}})Gupta, Runkler, Adel, Andrassy,
  Zimmermann, and Sch{\"u}tze}]{Gupta:89}
Pankaj Gupta, Thomas Runkler, Heike Adel, Bernt Andrassy, Hans-Georg
  Zimmermann, and Hinrich Sch{\"u}tze. 2015{\natexlab{a}}.
\newblock Deep learning methods for the extraction of relations in natural
  language text.
\newblock Technical report, Technical University of Munich, Germany.

\bibitem[{Gupta et~al.(2015{\natexlab{b}})Gupta, Runkler, and
  Andrassy}]{Gupta:86}
Pankaj Gupta, Thomas Runkler, and Bernt Andrassy. 2015{\natexlab{b}}.
\newblock Keyword learning for classifying requirements in tender documents.
\newblock Technical report, Technical University of Munich, Germany.

\bibitem[{Gupta et~al.(2016)Gupta, Sch{\"u}tze, and Andrassy}]{Gupta:90}
Pankaj Gupta, Hinrich Sch{\"u}tze, and Bernt Andrassy. 2016.
\newblock Table filling multi-task recurrent neural network for joint entity
  and relation extraction.
\newblock In {\em Proceedings of COLING 2016, the 26th International Conference
  on Computational Linguistics: Technical Papers\/}. Osaka, Japan, pages
  2537--2547.

\bibitem[{Gupta et~al.(2015{\natexlab{c}})Gupta, Sivalingam, P{\"o}lsterl, and
  Navab}]{Gupta:91}
Pankaj Gupta, Udhayaraj Sivalingam, Sebastian P{\"o}lsterl, and Nassir Navab.
  2015{\natexlab{c}}.
\newblock Identifying patients with diabetes using discriminative restricted
  boltzmann machines.
\newblock Technical report, Technical University of Munich, Germany.

\bibitem[{Hall et~al.(2008)Hall, Jurafsky, and Manning}]{hall:82}
David Hall, Daniel Jurafsky, and Christopher~D. Manning. 2008.
\newblock Studying the history of ideas using topic models.
\newblock In {\em Proceedings of the conference on Empirical Methods in Natural
  Language Processing\/}. Association for Computational Linguistics, Honolulu,
  Hawaii, pages 363--371.

\bibitem[{Hinton and Salakhutdinov(2009)}]{Rus:81}
Geoffrey Hinton and Ruslan Salakhutdinov. 2009.
\newblock Replicated softmax: an undirected topic model.
\newblock In {\em Advances in Neural Information Processing Systems 22\/}.
  Curran Associates, Inc., Vancouver, Canada, pages 1607--1614.

\bibitem[{Hinton(2002)}]{hin:82}
Geoffrey~E. Hinton. 2002.
\newblock Training products of experts by minimizing contrastive divergence.
\newblock {\em Neural Computation\/} 14(8):1771--1800.

\bibitem[{Hinton et~al.(2006)Hinton, Osindero, and Teh}]{hinton:82}
Geoffrey~E. Hinton, Simon Osindero, and Yee-Whye Teh. 2006.
\newblock A fast learning algorithm for deep belief nets.
\newblock {\em Neural Computation\/} 18(7):1527--1554.

\bibitem[{Iwata et~al.(2010)Iwata, Yamada, Sakurai, and Ueda}]{Iwata:82}
Tomoharu Iwata, Takeshi Yamada, Yasushi Sakurai, and Naonori Ueda. 2010.
\newblock Online multiscale dynamic topic models.
\newblock In {\em Proceedings of the 16th ACM SIGKDD International Conference
  on Knowledge Discovery and Data Mining\/}. Association for Computing
  Machinery, Washington DC, USA, pages 663--672.

\bibitem[{Newman et~al.(2009)Newman, Karimi, and Cavedon}]{Newman:82}
David Newman, Sarvnaz Karimi, and Lawrence Cavedon. 2009.
\newblock External evaluation of topic models.
\newblock In {\em Proceedings of the 14th Australasian Document Computing
  Symposium\/}. Citeseer, Sydney, Australia.

\bibitem[{Pruteanu-Malinici et~al.(2010)Pruteanu-Malinici, Ren, Paisley, Wang,
  and Carin}]{Malinici:82}
Iulian Pruteanu-Malinici, Lu~Ren, John Paisley, Eric Wang, and Lawrence Carin.
  2010.
\newblock Hierarchical bayesian modeling of topics in time-stamped documents.
\newblock {\em IEEE transactions on pattern analysis and machine
  intelligence\/} 32(6):996--1011.

\bibitem[{Rudolph and Blei(2018)}]{Rudolph:85}
Maja Rudolph and David Blei. 2018.
\newblock Dynamic bernoulli embeddings for language evolution.
\newblock In {\em Proceedings of the 27th International Conference on World
  Wide Web Companion\/}. Lyon, France.

\bibitem[{Rumelhart et~al.(1985)Rumelhart, Hinton, and Williams}]{rum:82}
David~E. Rumelhart, Geoffrey~E. Hinton, and Ronald~J. Williams. 1985.
\newblock Learning internal representations by error propagation.
\newblock Technical report, California Univ San Diego La Jolla Inst for
  Cognitive Science.

\bibitem[{Saha and Sindhwani(2012)}]{Saha:82}
Ankan Saha and Vikas Sindhwani. 2012.
\newblock Learning evolving and emerging topics in social media: a dynamic nmf
  approach with temporal regularization.
\newblock In {\em Proceedings of the 5th ACM International Conference on Web
  Search and Data Mining\/}. Association for Computing Machinery, Seattle,
  Washington USA, pages 693--702.

\bibitem[{Schein et~al.(2016)Schein, Wallach, and Zhou}]{Schein:82}
Aaron Schein, Hanna Wallach, and Mingyuan Zhou. 2016.
\newblock Poisson-gamma dynamical systems.
\newblock In {\em Advances in Neural Information Processing Systems 29\/},
  Curran Associates, Inc., Barcelona, Spain, pages 5005--5013.

\bibitem[{Smolensky(1986)}]{Smolensky:82}
Paul Smolensky. 1986.
\newblock Information processing in dynamical systems: Foundations of harmony
  theory.
\newblock Technical report, Colorado University at Boulder Department of
  Computer Science.

\bibitem[{Sutskever and Hinton(2007)}]{Sutskeverhinton:82}
Ilya Sutskever and Geoffrey Hinton. 2007.
\newblock Learning multilevel distributed representations for high-dimensional
  sequences.
\newblock In {\em Proceedings of the 11th International Conference on
  Artificial Intelligence and Statistics\/}. San Juan, Puerto Rico, pages
  548--555.

\bibitem[{Sutskever et~al.(2009)Sutskever, Hinton, and
  Taylor}]{Sutskeverhintontaylor:82}
Ilya Sutskever, Geoffrey~E. Hinton, and Graham~W. Taylor. 2009.
\newblock The recurrent temporal restricted boltzmann machine.
\newblock In {\em Advances in Neural Information Processing Systems 22\/}.
  Curran Associates, Inc., Vancouver, Canada, pages 1601--1608.

\bibitem[{Taylor et~al.(2007)Taylor, Hinton, and Roweis}]{taylorhinton:82}
Graham~W. Taylor, Geoffrey~E. Hinton, and Sam~T. Roweis. 2007.
\newblock Modeling human motion using binary latent variables.
\newblock In {\em Advances in Neural Information Processing Systems 20\/}.
  Curran Associates, Inc., Vancouver, Canada, pages 1345--1352.

\bibitem[{Vu et~al.(2016{\natexlab{a}})Vu, Adel, Gupta, and
  Sch{\"u}tze}]{Gupta:87}
Ngoc~Thang Vu, Heike Adel, Pankaj Gupta, and Hinrich Sch{\"u}tze.
  2016{\natexlab{a}}.
\newblock Combining recurrent and convolutional neural networks for relation
  classification.
\newblock In {\em Proceedings of the North American Chapter of the Association
  for Computational Linguistics: Human Language Technologies\/}. Association
  for Computational Linguistics, San Diego, California USA, pages 534--539.

\bibitem[{Vu et~al.(2016{\natexlab{b}})Vu, Gupta, Adel, and
  Sch{\"u}tze}]{Gupta:88}
Ngoc~Thang Vu, Pankaj Gupta, Heike Adel, and Hinrich Sch{\"u}tze.
  2016{\natexlab{b}}.
\newblock Bi-directional recurrent neural network with ranking loss for spoken
  language understanding.
\newblock In {\em Proceedings of the Acoustics, Speech and Signal Processing
  (ICASSP)\/}. IEEE, Shanghai, China, pages 6060--6064.

\bibitem[{Wan and Xiao(2008)}]{Wan:82}
Xiaojun Wan and Jianguo Xiao. 2008.
\newblock Single document keyphrase extraction using neighborhood knowledge.
\newblock In {\em Proceedings of the 23rd National Conference on Artificial
  Intelligence\/}. Chicago, Illinois USA, volume~8, pages 855--860.

\bibitem[{Wang and McCallum(2006)}]{xue:82}
Xuerui Wang and Andrew McCallum. 2006.
\newblock Topics over time: a non-markov continuous-time model of topical
  trends.
\newblock In {\em Proceedings of the 12th ACM SIGKDD International Conference
  on Knowledge Discovery and Data Mining\/}. Association for Computing
  Machinery, Philadelphia, Pennsylvania USA, pages 424--433.

\bibitem[{Xing et~al.(2005)Xing, Yan, and Hauptmann}]{Hauptmann:85}
Eric~P. Xing, Rong Yan, and Alexander~G. Hauptmann. 2005.
\newblock Mining associated text and images with dual-wing harmoniums.
\newblock In {\em Proceedings of the 21st Conference on Uncertainty in
  Artificial Intelligence\/}. AUAI Press, Edinburgh, Scotland UK.

\bibitem[{Yao et~al.(2018)Yao, Sun, Ding, Rao, and Xiong}]{Yao:85}
Zijun Yao, Yifan Sun, Weicong Ding, Nikhil Rao, and Hui Xiong. 2018.
\newblock Dynamic word embeddings for evolving semantic discovery.
\newblock In {\em Proceedings of the 11th ACM International Conference on Web
  Search and Data Mining (WSDM)\/}. Association for Computing Machinery, Los
  Angeles, California USA, pages 673--681.

\end{thebibliography}
\bibliographystyle{acl_natbib}

\end{document}